%% file: acl_latex.tex
\pdfoutput=1

\documentclass[11pt]{article}

\usepackage[preprint]{acl}

\usepackage{times}
\usepackage{latexsym}

\usepackage[T1]{fontenc}

\usepackage[utf8]{inputenc}

\usepackage{microtype}

\usepackage{inconsolata}

\usepackage{graphicx}

\title{What Do Vision–Language Models Encode for\\Personalized Image Aesthetics Assessment?}

\author{
  Koki Ryu${}^{1, 2}$ \quad
  Hitomi Yanaka${}^{1, 2, 3}$ \\
  ${}^1$The University of Tokyo \\
  ${}^2$Riken \\
  ${}^3$Tohoku University \\
  \texttt{\{kokiryu, hyanaka\}@is.s.u-tokyo.ac.jp}}

\usepackage{comment}
\usepackage{amsmath}
\usepackage{amsfonts}
\usepackage{booktabs}    %
\usepackage{tabularx}    %
\usepackage{xcolor}
\usepackage{subcaption}
\usepackage[most]{tcolorbox}
\usepackage{array}
\usepackage{siunitx}
\newenvironment{Prompt}[1][Prompt Example]{
  \begin{tcolorbox}[
    colback=white,
    colframe=black!50,
    boxrule=0.5pt,
    arc=2pt,
    outer arc=2pt,
    width=\linewidth,
    enlarge top by=5pt,
    enlarge bottom by=5pt
  ]
  \small
}{
  \end{tcolorbox}
}

\newcommand{\highlight}{\color{black} } %
\newcommand{\finishhighlight}{\color{black}}
\newcommand{\rhorcell}[2]{%
  \makebox[2.8em][r]{#1}%
  \hspace{0.25em}/\hspace{0.25em}%
  \makebox[2.8em][l]{#2}%
}
\newcommand{\cicell}[2]{$[#1,\ #2]$}

\begin{document}
\maketitle
\begin{abstract}
Personalized image aesthetics assessment (PIAA) is an important research problem with practical real-world applications.
While methods based on vision-language models (VLMs) are promising candidates for PIAA, it remains unclear whether they internally encode rich, multi-level aesthetic attributes required for effective personalization.
In this paper, we first analyze the internal representations of VLMs to examine the presence and distribution of such aesthetic attributes, and then leverage them for lightweight, individual-level personalization without model fine-tuning.
Our analysis reveals that VLMs encode diverse aesthetic attributes that propagate into the language decoder layers.
Building on these representations, we demonstrate that simple linear models can perform PIAA effectively.
We further analyze how aesthetic information is transferred across layers in different VLM architectures and across image domains.
Our findings provide insights into how VLMs can be utilized for modeling subjective, individual aesthetic preferences. \highlight Our code is available at \url{https://github.com/ynklab/vlm-latent-piaa}. \finishhighlight
\end{abstract}

\input{chapters/1_introduction.tex}
\input{chapters/2_related_work.tex}
\input{chapters/3_probing.tex}
\input{chapters/4_piaa.tex}
\input{chapters/5_conclusion.tex}

\newpage
\section*{Limitations}
Due to the difficulty of obtaining annotations for rich aesthetic attributes and personalized image aesthetics assessment, our experiments rely on a limited number of existing datasets.
Accordingly, the following limitations should be considered when interpreting our results:
\begin{itemize}

    \item The set of aesthetic attributes used in this study is not exhaustive.
    In addition, dataset-specific correlations among attributes may have influenced the linear probing results.
    \item For PIAA evaluation, we use a single dataset for each domain.
    As a result, trends interpreted as domain-specific may partly reflect dataset-specific biases, including those introduced by the annotator populations.
\end{itemize}

Regarding linear probing, it is important to note that the presence of linearly accessible information in hidden representations does not necessarily imply that VLMs directly utilize such information during text generation or scoring.
A more fine-grained analysis at the module or neuron level would be required to establish a closer connection between representation-level findings and model behavior at the output level.\par

Finally, our implementation of VLM-based PIAA baselines leaves room for improvement.
In particular, methods that directly interpret textual outputs as numeric scores discard information relevant to regression objectives, such as the relative proximity between scores.
More task-aligned optimization strategies could improve PIAA performance even for approaches that rely on text-based outputs.

\section*{Ethical Considerations}

Although the datasets used in this work include information related to annotator identity, such information is not utilized in our experiments.
Our analysis relies solely on subjective aesthetic scores associated with images, which poses minimal risk of personal data leakage.\par

Nevertheless, the proposed methods and findings could be combined with downstream recommendation or ranking systems that incorporate user attributes.
In such scenarios, any information that could enable personal identification should be handled with appropriate care and in accordance with relevant privacy regulations.\par

Additionally, personalization may introduce or amplify biases in aesthetics assessments for specific demographic groups.
As highlighted in prior studies, more comprehensive bias analyses are necessary before deploying personalized aesthetic assessment systems in real-world or industrial settings.

\highlight
\section*{Acknowledgments}
We thank the anonymous reviewers for their helpful comments and constructive feedback, and the meta-reviewer for summarizing the discussion and coordinating the review process.
We also thank Ryoma Kumon, Taisei Yamamoto and Tomoki Doi for their insightful discussions.
This work was supported by JSPS KAKENHI Grant Number JP24H00809, Japan.
\finishhighlight

\bibliography{custom}

\appendix
\input{chapters/appendix.tex}

\end{document}

%% file: chapters/1_introduction.tex
\section{Introduction}
\label{sec:introduction}
Image aesthetics assessment (IAA) is the task of evaluating the aesthetic quality of an input image.
Recently, personalized image aesthetics assessment (PIAA) has attracted increasing attention in the IAA field.
In this setting, models are trained to predict the aesthetics assessment that a specific user would assign to an image\highlight, thereby personalizing the predicted scores to reflect the user's aesthetic preferences\finishhighlight.
Several datasets for PIAA have been proposed~\cite{flicker-aes, para, lapis}, and they revealed substantial variation in aesthetic preferences across individuals.
Given practical applications such as social media platforms, it is necessary to align assessment models with individual preferences.\par

In PIAA, image-level aesthetic attributes such as lighting and color have been leveraged to better reflect individual preferences.
Many existing PIAA methods rely on training with large-scale, single-domain general image aesthetics assessment (GIAA) datasets~\cite{piaa-ici, piaa-mir, piaa-task-vector, flicker-aes} to extract such attributes from the target images.
However, such approaches require additional training costs.
Their transferability across different domains, such as photographs and artworks, is also questionable.\par
\input{figures/main-diagram}
To overcome these limitations, we propose a PIAA framework based on general vision-language models (VLMs).
For GIAA, prior studies have already employed VLMs to leverage their rich text-based data source~\cite{vila} or caption generation ability~\cite{image-captioning}.
However, their application to PIAA tasks has been limited to optimization at the demographic-group level rather than the individual level~\cite{aesbiasbench}.\par

In this work, we aim to achieve individual-level personalization by leveraging aesthetic attributes implicitly encoded in VLMs through large-scale pretraining.
Previous studies using linear probing have shown that VLM hidden representations capture semantic information~\cite{probing-semantic-rep, response-wide-shut} as well as overall image aesthetic scores~\cite{clip-knows-image-aesthetics}.
However, it remains unclear whether these representations also encode the diverse and continuous aesthetic attributes required for effective PIAA.
Therefore, we first conduct linear probing on the hidden layers of VLMs to verify the existence of rich aesthetic attributes.
We then apply linear regression to these representations and predict individual-level aesthetic scores, enabling personalized assessment without additional model fine-tuning.
We further analyze the domain-dependent behavior of our approach using datasets from two distinct domains: photographs and artworks.\par

Our contributions are summarized as follows:
\begin{itemize}
    \item We demonstrate that VLMs encode multiple aesthetic attributes beyond a single global aesthetic score within their hidden representations.
    Notably, this information propagates into the language decoder layers.
    Furthermore, we reveal that models with different architectures encode aesthetic attributes in different model regions.
    \item We show that simple linear regression on VLM hidden representations achieves strong PIAA performance, substantially outperforming methods based on text outputs such as few-shot prompting or fine-tuning.
    Our analysis indicates that the limited set of aesthetic attributes identified via linear probing plays a central role in personalization for photographs.
    Experiments on artwork datasets further suggest the presence of additional information in VLMs that contributes to PIAA but is not captured by probing based on photographs.
\end{itemize}

%% file: figures/main-diagram.tex
\begin{figure}[t]
  \centering
    \includegraphics[width=\linewidth]{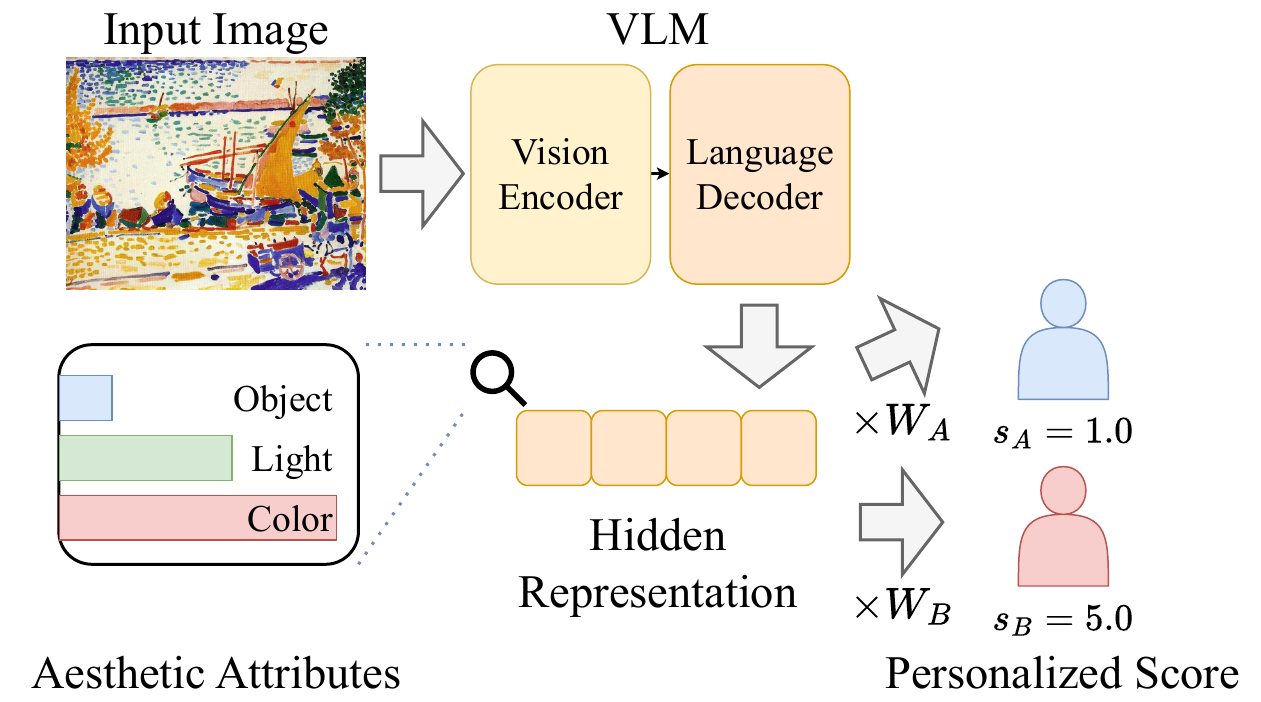}
  \caption{
  Overview of PIAA using VLM representations.
  $s$ and $W$ denote user-specific aesthetics scores and linear estimators respectively.
  Aesthetic attributes encoded in VLM hidden representations are linearly transformed to predict user-specific aesthetic scores without model fine-tuning.
  }
  \label{fig:main-diagram}
\end{figure}

%% file: chapters/2_related_work.tex
\section{Related Work}
\label{sec:related-work}

\subsection{Image Aesthetics Assessment}
\label{sec:related-work-iaa}
Several datasets have been proposed for general image aesthetics assessment (GIAA) over the past decade~\cite{ava, aadb, pccd}.
To extend GIAA to personalized image aesthetics assessment (PIAA), several datasets with user-specific annotations have been introduced.
FLICKER-AES~\cite{flicker-aes} was among the earliest datasets to include personalized aesthetic ratings.
PARA~\cite{para} and LAPIS~\cite{lapis} further expanded this line of work by providing richer image attribute annotations and user-specific ratings for photographs and artworks, respectively.\par

A variety of methods have also been proposed for PIAA.
\citet{piaa-meta} introduced a meta-learning approach to adapt models to individual users with limited annotations.
\citet{piaa-ici} and \citet{piaa-mir} modeled personalized aesthetics as interactions between image- and user-level attributes.
\citet{piaa-task-vector} proposed representing personalization as parameter changes induced by fine-tuning on general IAA tasks, referred to as a ``task vector,'' which is subsequently applied to user-specific prediction.\par

While these approaches have shown promising results, they typically require training on large-scale GIAA datasets followed by additional adaptation for each target user.
Such multi-stage pipelines incur substantial computational cost, and their cross-domain transferability (e.g., from photographs to artworks) remains unclear.

\subsection{Aesthetics Assessment with VLMs}
Several studies have explored the use of textual descriptions to train aesthetic assessment models based on VLM architectures.
\citet{vila} pretrained CoCa~\cite{coca} using aesthetics-related captions and demonstrated its effectiveness for downstream aesthetic assessment tasks.
\citet{uniqa} generated synthetic textual descriptions of image aesthetic attributes using VLMs and leveraged these data to train an aesthetics-aware image encoder.\par

Several benchmarks and fine-tuning approaches have also been proposed to evaluate and improve VLMs for aesthetic assessment~\cite{q-bench, uniaa, aesbench, q-instruct, photoeye}.
More recently, \citet{aesbiasbench} analyzed VLM behavior on PIAA datasets with respect to user attributes such as gender and age, revealing tendencies to over-align with specific demographic groups.\par
However, existing evaluations of VLMs' ability to perceive fine-grained aesthetic attributes are primarily limited to multiple-choice or binary formats.
As a result, it remains unclear whether VLMs encode continuous, multi-level aesthetic attributes required for personalized image aesthetics assessment.
Furthermore, to the best of our knowledge, no prior work has investigated individual-level PIAA using VLMs as the backbone model.

\subsection{Representation Analysis of VLMs}
Linear probing has been widely adopted as a tool for analyzing the internal representations of foundation models across various visual tasks~\cite{probe-dino, probe-vit}.
\citet{clip-knows-image-aesthetics} applied linear probing to CLIP~\cite{clip} to evaluate its understanding of general image aesthetics.\par

Another line of work emphasizes the importance of integrated analyses spanning multiple transformer layers across both vision encoders and language decoders.
\citet{eyes-wide-shut} demonstrated that certain visual attributes are challenging for vision encoders trained with contrastive image-text objectives to capture.
In contrast, \citet{response-wide-shut} demonstrated that information relevant to fine-grained recognition tends to diminish in language decoder layers while remaining more stable in vision encoders.
\citet{probing-semantic-rep} conducted layer-wise probing of language decoders, suggesting that different layers encode different types of information.
Several studies have further analyzed how visual information is transferred from vision tokens to text tokens through the language decoder~\cite{whats-in-the-image, cross-modal-info}.\par

Despite these insights, most existing analyses involving language decoders focus on a limited set of tasks, such as object recognition or visual question answering.
Although \citet{clip-knows-image-aesthetics} examined aesthetic understanding at the representation level, it remains unclear how multiple, fine-grained aesthetic attributes are encoded and propagated across different layers of VLMs.
Importantly, our probing is designed not merely to assess overall aesthetic awareness, but to reveal multi-attribute representations that enable personalization.

%% file: chapters/3_probing.tex
\section{Probing Aesthetic Attributes in VLMs}
\label{sec:probing}
In this section, we investigate whether VLMs encode rich, multi-level aesthetic attributes that are relevant to PIAA.

\subsection{Method Overview}
We perform linear probing on the internal representations of VLMs to quantify the extent to which aesthetic attribute information is encoded.
Formally, let $I$ denote an input image, $\mathbf{v}_I \in \mathbb{R}^K$ the $K$-dimensional ground-truth aesthetic attribute vector (e.g., object, lighting, and color), and $\mathbf{h}(I) \in \mathbb{R}^D$ the $D$-dimensional hidden representation extracted from a VLM for image $I$.
Our objective is to learn an image-agnostic linear transformation $M \in \mathbb{R}^{K \times D}$ such that
\begin{align}
    M \mathbf{h}(I) \approx \mathbf{v}_I .
\end{align}

In practice, we estimate $M$ using ridge regression,
which performs a stable estimation for high-dimensional representations while mitigating overfitting through L2 regularization.
Detailed implementation is provided in Appendix~\ref{appendix:probing-implementation}.
For the hidden representation $\mathbf{h}(I)$, we extract output hidden vectors from each transformer layer of the VLM, using the image together with the prompt \textit{``Assess the aesthetics of this image.''} as input.
To account for potential prompt sensitivity, we examine the robustness of probing results to alternative prompt formulations in Appendix~\ref{appendix:prompt-sensitivity}.
\par

We use average pooling to obtain a single representation for each transformer layer, as visual information in VLMs is distributed across tokens and no dedicated image-level token exists.
This provides a simple, modality-agnostic aggregation that enables fair comparison across vision and language representations.\par

We consider three layer-wise representations obtained via average pooling: $\mathbf{V}_i$ (vision encoder), $\mathbf{LT}_i$ (language decoder, text tokens), and $\mathbf{LV}_i$ (language decoder, vision tokens).
We primarily focus on comparisons involving $\mathbf{LT}_i$, since aesthetic attribute encoding in language decoder layers remains less explored in prior work~\cite{clip-knows-image-aesthetics} and $\mathbf{LT}_i$ representations directly support text-based outputs such as aesthetic scores and captions.
For completeness, we also compare $\mathbf{LT}_i$ with the last text-token representation in the language decoder in Appendix~\ref{appendix:full-aadb-probing}.

\subsection{Settings}
\label{sec:probing-settings}

\paragraph{Datasets}
We use two photographic datasets for the probing experiments.
Our primary dataset is AADB~\cite{aadb}, which provides 11-dimensional aesthetic attribute annotations alongside overall aesthetic scores, with continuous values in the range $[-1, 1]$.
We also conduct probing experiments on PARA~\cite{para}, which includes several general aesthetic attribute annotations.
For each hidden representation, we train attribute regressors on the training split and report evaluation metrics on the test split.
\par
Note that the attribute annotations in PARA exhibit strong inter-attribute correlations as shown in Appendix~\ref{appendix:dataset-stats}.
Such correlations are not well aligned with our goal of identifying diverse and independent aesthetic attributes encoded in VLM representations.
Therefore, we report results on AADB, where inter-attribute correlations are relatively low, as the primary probing results in the following section, and present PARA-based probing results in Appendix~\ref{appendix:probing-on-para} as supplementary material.

\paragraph{Models}
We evaluate two state-of-the-art open-source VLMs with distinct architectures: Qwen3-VL~\cite{qwen3vl} and Gemma~3~\cite{gemma3}.
Gemma~3 uses a fixed number of visual tokens and feeds only the final vision encoder output into the language decoder, whereas Qwen3-VL produces resolution-dependent vision tokens and integrates multi-level vision representations via DeepStack~\cite{deepstack}.
All models are instruction-tuned variants, and we evaluate multiple model sizes (2B, 4B, and 8B for Qwen3-VL; 4B and 12B for Gemma~3).\par
We additionally include DINOv3 (ViT-B/16, ViT-L/16)~\cite{dinov3} as a vision-only foundation model for comparison.

\paragraph{Evaluation Methods}
We use Spearman's rank correlation coefficient as the primary evaluation metric.
For each model, we first report the best correlation obtained across different $\mathbf{LT}_i$ layers (or $\mathbf{V}_i$ layers for DINOv3).
We then compare these values with the corresponding results from $\mathbf{LV}_i$ and $\mathbf{V}_i$ representations.

\subsection{Results}
\label{section:probing-result}
\input{tables/probing_table}
\input{figures/probing}

\input{figures/probing_attr}
\paragraph{Overall Results}
The main results on $\mathbf{LT}$ layers (and $\mathbf{V}$ layers for DINOv3) are summarized in Table~\ref{tab:probing_best_layers}.
Results for the other representations are reported in Appendix~\ref{appendix:full-aadb-probing}.
Layer-wise results for all $\mathbf{V}$ and $\mathbf{LT}$ layers of Qwen3-VL~2B are further visualized in Figure~\ref{fig:probing}.\par

Across more than half of the aesthetic attributes, all VLMs achieve moderate positive correlations (greater than 0.4).
For the remaining attributes, which exhibit sparse label distributions (see Appendix~\ref{appendix:dataset-stats}), the models still achieve consistently positive correlations.
We observe similar trends across alternative prompt formulations reported in Appendix~\ref{appendix:prompt-sensitivity}.
These results indicate that VLMs encode a diverse set of aesthetic attributes in their hidden representations in a manner that is linearly accessible.
To address the possibility that correlated attributes or spurious visual cues drive these effects, we conduct additional robustness analyses based on controlled image augmentations, which are reported in Appendix~\ref{appendix:probing-aug}.
\par 

When comparing different aesthetic attributes, probing performance for the overall aesthetic score is consistently higher than that for most fine-grained attributes.
This observation suggests that the ability to capture general aesthetics, previously verified through the probing study by \citet{clip-knows-image-aesthetics}, does not necessarily imply a robust encoding of fine-grained aesthetic attributes.\par

\paragraph{Comparison of Models and Components}
Interestingly, Qwen3-VL~2B, 4B, and 8B achieve the best performance for different attributes, respectively.
The fact that smaller models within the same model family can outperform larger variants suggests that aesthetic attribute encoding is not directly correlated with conventional VLM benchmark performance, such as visual question answering.\par

Another notable observation from Figure~\ref{fig:probing} is that language decoder representations achieve higher correlations for a larger number of attributes than vision encoder representations.
While vision encoder layers achieve better performance for specific attributes (e.g., \textit{VividColor} in Qwen3-VL~2B), no substantial performance degradation is observed in the language decoder layers.
Moreover, as Table~\ref{tab:probing_best_layers} shows, DINOv3 consistently yields the lowest correlations across nearly all attributes.
Together, these findings suggest that language decoder layers play an important role in encoding aesthetic attribute information beyond what is captured by vision-only foundation models.\par

\paragraph{Layer-wise Analysis}
We further analyze layer-wise probing performance for the \textit{Content} attribute across $\mathbf{V}$, $\mathbf{LT}$, and $\mathbf{LV}$ representations in Figure~\ref{fig:probing-attr}.
For Gemma~3, performance in $\mathbf{LT}$ representations improves notably in the early to middle layers of the language decoder as layer depth increases.
This observation is consistent with prior studies~\cite{whats-in-the-image, cross-modal-info}, which report that visual information relevant to textual outputs is transferred to text tokens in the lower to middle layers of the language decoder.
In contrast, this trend is not observed for Qwen3-VL, where probing performance for $\mathbf{LT}$ and $\mathbf{LV}$ representations remains comparable across layers.
This pattern is consistently observed across multiple attributes.\par
We hypothesize that this architectural difference stems from the way aesthetic information is integrated across different modalities.
Specifically, Gemma~3 may process aesthetics information primarily within the language decoder, exhibiting its potential dependence on text supervision. At the same time, Qwen3-VL may encode a larger part of such information within the vision encoder due to its DeepStack architecture.

%% file: tables/probing_table.tex
\begin{table*}[t]
\centering
\small
\setlength{\tabcolsep}{3pt}
\renewcommand{\arraystretch}{1.0}
\begin{tabular}{
l
S @{\,(} c @{)\hspace{1em}}
S @{\,(} c @{)\hspace{1em}}
S @{\,(} c @{)\hspace{1em}}
S @{\,(} c @{)\hspace{1em}}
S @{\,(} c @{)\hspace{1em}}
S @{\,(} c @{)\hspace{1em}}
S @{\,(} c @{)}
}
\toprule
Attribute
& \multicolumn{6}{c}{Qwen3-VL}
& \multicolumn{4}{c}{Gemma 3}
& \multicolumn{4}{c}{DINOv3} \\
& \multicolumn{2}{c}{2B}
& \multicolumn{2}{c}{4B}
& \multicolumn{2}{c}{8B}
& \multicolumn{2}{c}{4B}
& \multicolumn{2}{c}{12B}
& \multicolumn{2}{c}{ViT-B/16}
& \multicolumn{2}{c}{ViT-L/16} \\
\midrule
BalancingElements & 0.325 & 13 & \bfseries 0.332 & 0 & 0.300 & 14 & 0.309 & 24 & 0.307 & 14 & 0.317 & 11 & 0.307 & 22 \\
ColorHarmony      & 0.516 & 9  & \bfseries 0.523 & 9 & 0.515 & 6  & 0.493 & 24 & 0.504 & 32 & 0.479 & 6  & 0.482 & 10 \\
Content           & \bfseries 0.633 & 10 & 0.632 & 13 & 0.627 & 15 & 0.621 & 12 & 0.624 & 35 & 0.551 & 12 & 0.579 & 17 \\
DoF               & \bfseries 0.535 & 10 & 0.518 & 10 & 0.530 & 16 & 0.512 & 9  & 0.515 & 15 & 0.506 & 7  & 0.507 & 18 \\
Light             & \bfseries 0.509 & 14 & 0.507 & 12 & 0.490 & 14 & 0.452 & 18 & 0.468 & 6  & 0.439 & 8  & 0.436 & 11 \\
MotionBlur        & 0.165 & 12 & 0.134 & 5  & \bfseries 0.188 & 9 & 0.155 & 1  & 0.152 & 37 & 0.161 & 7  & 0.143 & 3 \\
Object            & \bfseries 0.722 & 18 & 0.719 & 19 & 0.716 & 16 & 0.706 & 18 & 0.714 & 7  & 0.688 & 12 & 0.696 & 19 \\
Repetition        & \bfseries 0.461 & 3  & 0.446 & 14 & 0.451 & 4  & 0.415 & 8  & 0.430 & 20 & 0.438 & 8  & 0.451 & 19 \\
RuleOfThirds      & \bfseries 0.288 & 11 & 0.267 & 5  & 0.266 & 0  & 0.267 & 12 & 0.273 & 15 & 0.230 & 9  & 0.230 & 20 \\
Symmetry          & 0.315 & 10 & \bfseries 0.329 & 14 & 0.307 & 6  & 0.281 & 11 & 0.302 & 33 & 0.299 & 5  & 0.313 & 11 \\
VividColor        & 0.686 & 0  & 0.695 & 11 & \bfseries 0.696 & 0 & 0.671 & 10 & 0.687 & 18 & 0.686 & 3  & 0.685 & 10 \\
Overall Score     & 0.725 & 5  & \bfseries 0.727 & 19 & 0.720 & 10 & 0.700 & 10 & 0.719 & 13 & 0.636 & 9  & 0.666 & 17 \\
\bottomrule
\end{tabular}
\caption{Highest Spearman correlation achieved by linear probing on $\mathbf{LT}$ layers for each aesthetic attribute in AADB.
Values in parentheses indicate the corresponding layer indices.}
\label{tab:probing_best_layers}
\end{table*}

%% file: figures/probing.tex
\begin{figure}[t]
  \centering
    \includegraphics[width=\columnwidth]{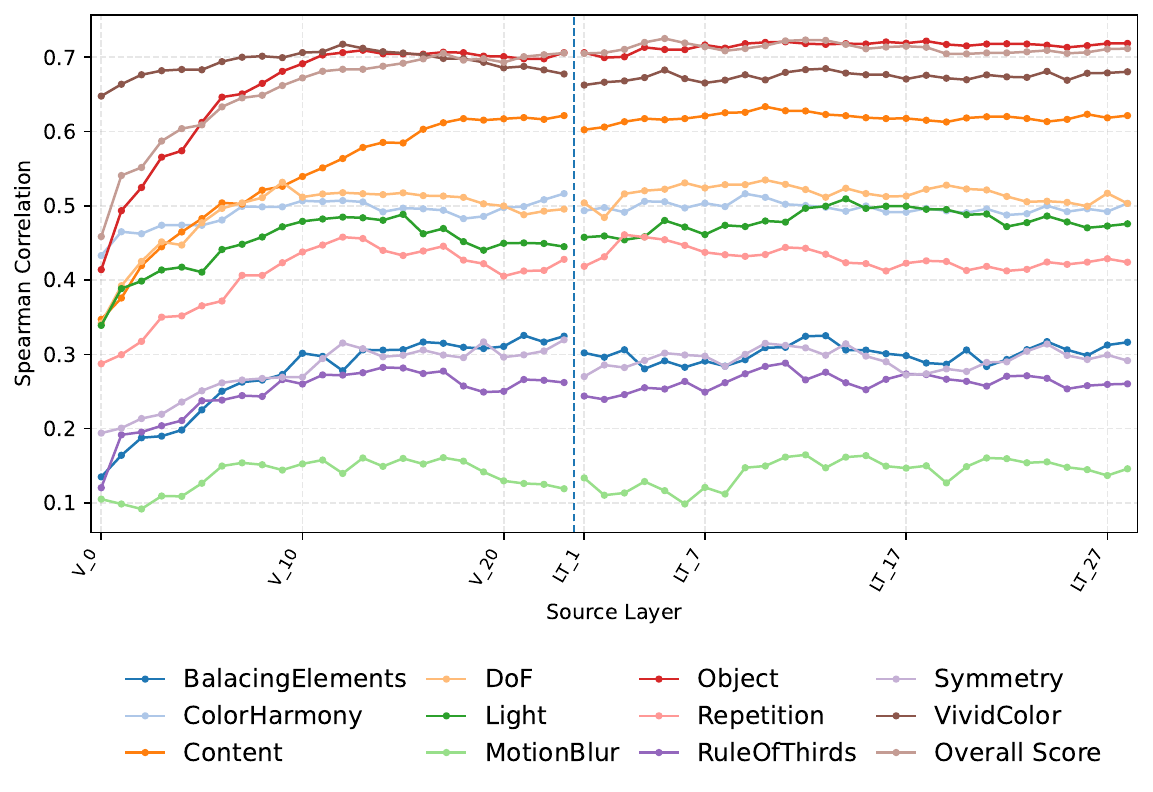}
  \caption{Layer-wise probing performance across $\mathbf{V}$ and $\mathbf{LT}$ layers for Qwen3-VL 2B on AADB.
The dotted line indicates the boundary between $V$ and $LT$.}
  \label{fig:probing}
\end{figure}

%% file: figures/probing_attr.tex
\begin{figure}[t]
  \centering
  \begin{minipage}[t]{0.5\linewidth}
    \centering
    \includegraphics[width=\linewidth]{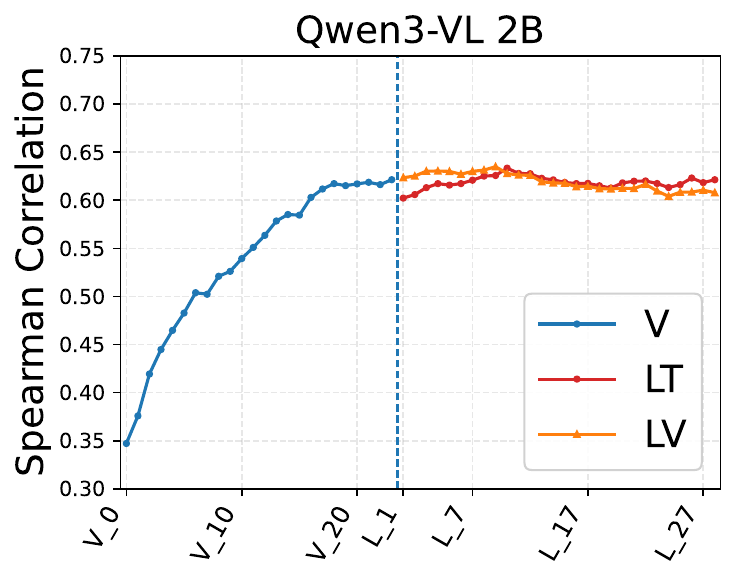}
  \end{minipage}%
  \begin{minipage}[t]{0.5\linewidth}
    \centering
    \includegraphics[width=\linewidth]{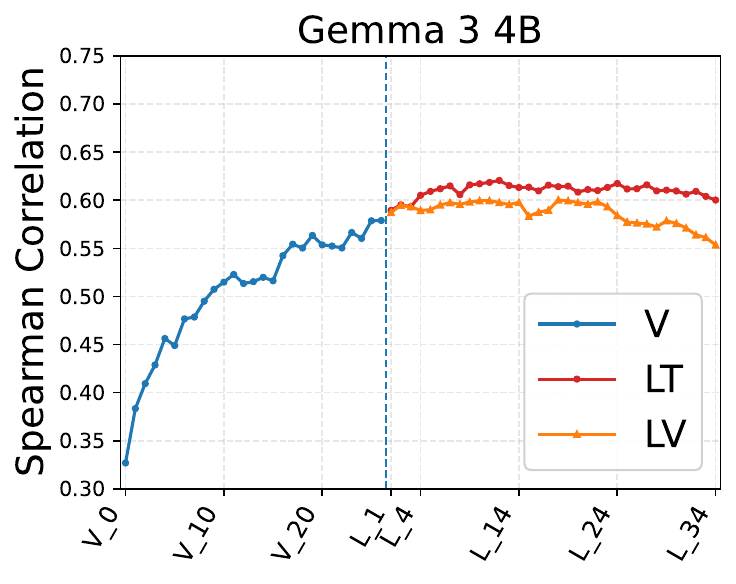}
  \end{minipage}%

  \caption{Layer-wise probing performance for the \textit{Content} attribute in Qwen3-VL 2B and Gemma~3 4B.}
  \label{fig:probing-attr}
\end{figure}

%% file: chapters/4_piaa.tex
\section{PIAA with VLMs}
\label{sec:piaa}
In this section, we perform PIAA using hidden representations of VLMs, leveraging the rich aesthetic attributes identified in Section~\ref{sec:probing}.

\subsection{Method Overview}
We aim to predict personalized aesthetic scores, which may vary across users for the same image.\par
Formally, let $u$ denote a target user, $I$ an image, $\mathbf{h}(I)$ a hidden representation extracted from a VLM, and $s_{I,u}$ the personalized aesthetic score assigned by user $u$ to image $I$.
Our objective is to learn a user-specific, image-agnostic linear transformation $M_u$ such that
\begin{align}
    M_u \mathbf{h}(I) \approx s_{I,u} .
\end{align}
We train $M_u$ using user-specific training data and evaluate its performance on held-out test data for the same user.\par
As in Section~\ref{sec:probing}, we estimate $M_u$ via ridge regression.
For $\mathbf{h}(I)$, we reuse the $\mathbf{V}_i$, $\mathbf{LT}_i$, and $\mathbf{LV}_i$ representations defined in Section~\ref{sec:probing}.
Based on the probing analysis in Section~\ref{sec:probing}, we observe that the layer at which aesthetic attribute information peaks varies across models and attributes. In contrast, language decoder representations in the middle layers consistently contain substantial information.
Accordingly, we use $\mathbf{LT}_{15}$ as a representative layer for reporting the main PIAA results, as it provides a stable and informative representation across different models.
\highlight We also report results using $\mathbf{V}_i$ representations in Appendix~\ref{appendix:vision-encoder-feature-piaa}, given that prior work~\cite{clip-knows-image-aesthetics} has explored GIAA using vision-encoder representations. \finishhighlight

The same prompt used for representation extraction in Section~\ref{sec:probing} is also adopted here.
We refer to this primary approach as \textbf{Linear-Hidden}.

\subsection{Evaluation Settings}
\label{sec:piaa-other-settings}
We conduct experiments on two PIAA datasets from different domains introduced in Section~\ref{sec:related-work-iaa}: PARA~\cite{para}, which consists of photographs, and LAPIS~\cite{lapis}, which focuses on artworks.
For each dataset, we randomly sample 200 users.
For each user, we construct a personalized support set with either 10 images (small setting) or 100 images (large setting), and reserve 50 images as a personalized test set.
Unless otherwise specified, we report results under the 100-shot setting.
Full results for both 10-shot and 100-shot settings are provided in Appendix~\ref{appendix:full-piaa-results}.
Since LAPIS annotations are provided on a $[0,100]$ scale whereas PARA uses a $[1,5]$ scale, we linearly rescale LAPIS annotations to the $[1,5]$ range prior to training and evaluation.\par
As target models, we use the same Qwen3-VL and Gemma~3 models evaluated in Section~\ref{sec:probing}.\par
While prior work on PARA~\cite{para} reports correlation values aggregated over all test subjects as the primary evaluation metric, we observe that such metrics can be artificially improved through simple per-user, image-agnostic numeric adjustments applied uniformly across all images.
To disentangle this numeric calibration effect from genuine user-specific preference modeling, we evaluate performance at the individual user level.
Specifically, we compute Spearman’s rank correlation coefficient ($\rho$) and the coefficient of determination ($R^2$) separately for each user to capture complementary aspects of personalization performance: $\rho$ measures the consistency of relative preference ordering. At the same time, $R^2$ reflects the accuracy of absolute score prediction.
We then report the user-averaged metrics across users as our main evaluation results.
\highlight Spearman’s rank correlation becomes undefined for a user when all predicted values are identical.
In such cases, we exclude the user from the averaging process.
We validate this experimental design in Appendix~\ref{appendix:validate-undefined-correlations}.
\finishhighlight
\highlight
We further assess the statistical robustness of our results using bootstrap resampling, as described in Appendix~\ref{sec:statistical-validation}.
\finishhighlight

\subsection{VLM-based Baselines}

We compare Linear-Hidden against several VLM-based baselines designed to evaluate different aspects of personalization.\par

To better understand the source of personalization effects, we first consider two variants of the Linear-Hidden model.
In \textbf{Linear-Hidden (GIAA)}, we replace personalized scores with non-user-specific GIAA scores as regression targets.
This setting isolates general aesthetic perception from user-specific preference modeling.\par

In \textbf{Linear-Hidden (Reduce)}, we first train a user-agnostic regressor $M$ on AADB to predict aesthetic attributes excluding the overall score, and then train a user-specific regressor $M'_u$ such that
\begin{align}
    M'_u (M \mathbf{h}(I)) \approx s_{I,u} .
\end{align}
Since the intermediate transformation $M$ substantially reduces representation dimensionality, this variant evaluates whether the aesthetic attributes identified in Section~\ref{sec:probing} are sufficient to support PIAA prediction.\par

In addition, we include text-based baselines that do not directly access hidden representations.
We prompt VLMs to output GIAA scores as text without user-specific conditioning (\textbf{Raw Text}), as well as conventional adaptation methods (\textbf{Few-shot} and \textbf{LoRA}).
\highlight Due to the high memory cost of long-context prompting, we use the 10-shot setting for the \textbf{Few-shot} baseline. \finishhighlight

We also include the \textbf{Adjust-Bias} baseline, which applies a user-specific additive bias correction to \textbf{Raw Text} predictions based on training-set errors, allowing us to disentangle score calibration from preference ordering.
Implementation details of all baselines are provided in Appendix~\ref{appendix:piaa-baseline-details}.
\highlight

\subsection{Domain-Specific Baseline}

We further include a domain-specific PIAA baseline for comparison with Linear-Hidden.
We choose PIAA-ICI~\cite{piaa-ici}, a recent dedicated PIAA model that has demonstrated strong performance on both PARA and LAPIS.

\paragraph{In-Domain Comparison on PARA}

We first train PIAA-ICI solely on the PARA dataset to evaluate its performance in an in-domain setting.

PIAA-ICI consists of two stages: user-agnostic pretraining and user-specific fine-tuning.
For pretraining, we use 238 users and 12,204 images from PARA, ensuring that neither the users nor the images overlap with those used in the evaluation.

The original PIAA-ICI model incorporates both image attributes (e.g., color and lighting) and user attributes (e.g., age and gender).
Since Linear-Hidden does not utilize user attributes, we adopt the variant without user attributes (denoted as \textit{w/o User Attr}) as our primary domain-specific baseline.
For completeness, we also report results using the full model with user attributes (denoted as \textit{w/ User Attr}).
The pretraining set includes 7 image attributes and 5 user attributes based on the annotations available in PARA.

After pretraining, we perform personalized fine-tuning using 100 images per user and evaluate performance on 50 held-out images per user, consistent with our main experimental protocol.

\paragraph{Cross-Domain Transfer to LAPIS}
We further evaluate the domain transferability of PIAA-ICI by separating the data used for pretraining and fine-tuning.
Specifically, we apply the model pretrained on PARA to LAPIS for personalized fine-tuning and evaluation, without any additional pretraining on LAPIS.
Since PARA and LAPIS do not share common user attributes, we use the variant without user attribute inputs for cross-domain evaluation.
For both methods, we use 100 in-domain LAPIS images per user for fine-tuning before evaluation.
\finishhighlight

\subsection{Results}
\label{sec:piaa-results}
\input{tables/concat_latex_table}

\input{figures/layerwise_piaa}

\input{figures/repr_user}
\input{tables/piaa_ici_in_domain}
\input{tables/piaa_ici_cross_domain}
\input{tables/hard_latex_table}
\paragraph{Overall Results}
Table~\ref{tab:para_lapis_useravg_metrics} summarizes the main PIAA results \highlight with VLM-based baselines\finishhighlight.
Complete results for different support set sizes are provided in Appendix~\ref{appendix:full-piaa-results}.\par

Across all models and both datasets, the Linear-Hidden approach consistently outperforms text-based baselines, including Raw Text, Few-shot prompting, and LoRA fine-tuning.
Notably, on the LAPIS dataset, while text-based baselines such as Raw Text and Few-shot prompting already exhibit low performance, LoRA performs even worse.
One possible explanation is the difficulty of learning fine-grained image–score relationships using token-level likelihood objectives under limited data.
Despite this difficulty, Linear-Hidden achieves high Spearman correlations (above 0.5).
These results indicate that language decoder representations in VLMs encode sufficiently rich information for image aesthetics assessment.\par

\paragraph{Domain Comparison}
Results obtained with Linear-Hidden variants reveal domain-specific characteristics.
While Linear-Hidden (GIAA) yields substantially worse $R^2$ values than the full PIAA setting, the two settings exhibit only minor differences in Spearman correlation.
This suggests that general aesthetics scores primarily contribute to image-agnostic numerical calibration across users, but do not fully capture individual preference ordering.

In contrast, training with user-specific PIAA labels results in significant improvements in both Spearman correlation and $R^2$ on the LAPIS dataset.
Figure~\ref{fig:repr_user} illustrates qualitative differences between GIAA-based and PIAA-based predictions for a representative user in LAPIS.
While the GIAA-based model favors fine-grained drawings, the PIAA-based model more accurately reflects the user’s preference for colorful and abstract artworks, as well as their disinterest in realistic human portraits.\par

Comparisons with Linear-Hidden (Reduce) further highlight domain-specific behavior.
On PARA, the Reduced variant achieves performance comparable to the complete Linear-Hidden model.
However, on LAPIS, a clear performance gap exists between the two methods.
This observation suggests that the 11-dimensional aesthetic attribute space identified in Section~\ref{sec:probing} is sufficient for PIAA in photographs, whereas personalization in the artwork domain relies on additional attributes that are also encoded in VLM representations but are not captured by the probed attributes.\par

\paragraph{Layer-wise Analysis}
Layer-wise comparisons across $\mathbf{V}$, $\mathbf{LV}$, and $\mathbf{LT}$ representations are shown in Figure~\ref{fig:layerwise_piaa}.
For PARA, both Qwen3-VL and Gemma~3 exhibit peak performance in the middle layers of $\mathbf{LT}$, with $\mathbf{LT}$ consistently outperforming $\mathbf{V}$ and $\mathbf{LV}$ across layers.
In contrast, this trend does not hold for LAPIS.
Moreover, for Qwen3-VL, $\mathbf{LV}$ representations consistently outperform $\mathbf{LT}$ across layers on LAPIS.\par

We hypothesize that these differences are due to the domain-specific availability of aesthetics-related textual supervision.
While photographs benefit from rich caption and critique datasets containing aesthetic content~\cite{ava-captions, photoeye, pccd}, comparable resources are scarce for artworks.
As a result, instruction-tuned VLMs may integrate aesthetic information into text tokens for photographs, but rely more heavily on vision-side representations for artworks.

\highlight
In Appendix~\ref{sec:piaa-pair}, we further evaluate PIAA using combined representations from \textbf{V} and the \textbf{LT}, and find that these representations provide complementary information, leading to consistent performance improvements.

\paragraph{Comparison with the Domain-Specific Baseline}
Tables~\ref{tab:piaa_ici_in_domain} and \ref{tab:piaa_ici_cross_domain} compare Linear-Hidden with PIAA-ICI under in-domain and cross-domain settings, respectively.
In the PARA-only in-domain setting, Linear-Hidden achieves performance comparable to PIAA-ICI, despite being trained solely on the 100-shot personal support set without attribute supervision.
While incorporating user attributes improves absolute score calibration (reflected in $R^2$), it has a limited impact on ranking performance ($\rho$).

In the cross-domain setting, Linear-Hidden significantly outperforms PIAA-ICI.
This suggests that domain-specific pretraining on a single dataset may limit cross-domain generalization in dedicated methods, whereas VLM hidden representations exhibit stronger robustness under domain shift.
Motivated by this observation, we further investigate a more challenging domain generalization setting in Appendix~\ref{sec:piaa-domain-transfer}, where no in-domain images are available even in the user-specific support sets.
\finishhighlight

\subsection{Additional Analysis on ``Hard'' Users}
\label{sec:piaa-additional-analysis}

\input{figures/annotator_agreement}
We observe that the personalization effect of Linear-Hidden is weaker on PARA than on LAPIS.
However, this difference may partly be because of the label distribution characteristics.
As shown in Figure~\ref{fig:annotator_agreement}, annotator preferences in PARA exhibit substantially higher agreement with GIAA scores compared to those in LAPIS.
This indicates that personalization is fundamentally more challenging on PARA, as user-specific variations from general aesthetics are limited.
To isolate this effect from the domain difference, we resample 50 users from PARA with the lowest Spearman correlation between their annotations and GIAA scores, and repeat the PIAA experiments.\par

Results for these ``hard'' users are presented in Table~\ref{tab:hard_dataset_useravg_metrics}.
Although the performance gap remains smaller than that observed on LAPIS, regressors trained with PIAA labels show clear improvements over GIAA-based baselines.
This finding demonstrates that VLM-based personalization is feasible for photographs when annotator preferences are sufficiently distinct from one another.\par

Notably, under this setting, the Reduced variant consistently underperforms the full Linear-Hidden in terms of Spearman correlation.
This suggests that personalization for specific preferences in photographs requires additional attributes beyond those identified through our linear probing.
At the same time, the Reduced variant exhibits improved $R^2$ scores compared to the full Linear-Hidden, which may be attributed to enhanced numerical stability resulting from the reduced number of explanatory variables.

%% file: tables/concat_latex_table.tex
\begin{table*}[t]
\centering
\small
\setlength{\tabcolsep}{3pt}
\renewcommand{\arraystretch}{1.0}

\begin{tabular}{
l l
*{5}{S[table-format=1.3] @{\,/\,} S[table-format=1.3,table-format=-1.3]}
}
\toprule
Method & Support
& \multicolumn{6}{c}{Qwen3-VL}
& \multicolumn{4}{c}{Gemma 3} \\

& 
& \multicolumn{2}{c}{2B}
& \multicolumn{2}{c}{4B}
& \multicolumn{2}{c}{8B}
& \multicolumn{2}{c}{4B}
& \multicolumn{2}{c}{12B} \\

&
& \multicolumn{2}{c}{$\rho$ / $R^2$}
& \multicolumn{2}{c}{$\rho$ / $R^2$}
& \multicolumn{2}{c}{$\rho$ / $R^2$}
& \multicolumn{2}{c}{$\rho$ / $R^2$}
& \multicolumn{2}{c}{$\rho$ / $R^2$} \\
\midrule

\multicolumn{12}{l}{\textbf{PARA}} \\
\midrule

Raw Text &
& 0.504 & -0.571 & 0.570 & -1.277 & 0.528 & -0.729 & 0.462 & -1.107 & 0.493 & -1.879 \\

Few-shot & 10-shot
& 0.319 & -1.850 & 0.197 & -1.576 & 0.372 & -0.547 & 0.241 & -0.537 & 0.407 & -0.185 \\

Adjust-Bias & 100-shot
& 0.504 & -0.310 & 0.570 & -0.672 & 0.528 & -0.441 & 0.462 & -0.321 & 0.493 & -1.562 \\

LoRA & 100-shot
& 0.487 & -1.970 & 0.578 & -1.751 & 0.568 & -0.978 & 0.489 & -0.893 & 0.524 & -0.525 \\

Linear-Hidden & 100-shot
& \bfseries 0.604 & 0.363
& \bfseries 0.611 & 0.362
& 0.591 & \bfseries 0.341
& 0.591 & 0.346
& 0.594 & 0.329 \\

Linear-Hidden (GIAA) & 100-shot
& 0.596 & 0.041
& 0.603 & 0.057
& \bfseries 0.596 & 0.043
& 0.584 & -0.014
& \bfseries 0.594 & 0.036 \\

Linear-Hidden (Reduce) & 100-shot
& 0.585 & \bfseries 0.367
& 0.597 & \bfseries 0.382
& 0.558 & 0.322
& \bfseries 0.592 & \bfseries 0.365
& 0.593 & \bfseries 0.373 \\

\midrule\midrule

\multicolumn{12}{l}{\textbf{LAPIS}} \\
\midrule

Raw Text &
& 0.098 & -0.778 & 0.176 & -0.937 & 0.175 & -0.763 & 0.119 & -1.340 & 0.233 & -1.335 \\

Few-shot & 10-shot
& 0.142 & -1.265 & 0.221 & -0.380 & 0.264 & -0.480 & 0.127 & -0.354 & 0.227 & -0.459 \\

Adjust-Bias & 100-shot
& 0.098 & -0.264 & 0.176 & -0.231 & 0.175 & -0.206 & 0.119 & -0.162 & 0.233 & -0.442 \\

LoRA & 100-shot
& 0.026 & -0.701 & 0.153 & -1.580 & 0.164 & -1.386 & 0.116 & -0.936 & 0.201 & -1.022 \\

Linear-Hidden & 100-shot
& \bfseries 0.568 & \bfseries 0.321
& \bfseries 0.568 & \bfseries 0.319
& \bfseries 0.573 & \bfseries 0.313
& \bfseries 0.568 & \bfseries 0.328
& \bfseries 0.571 & \bfseries 0.323 \\

Linear-Hidden (GIAA) & 100-shot
& 0.418 & -0.148 & 0.420 & -0.148 & 0.420 & -0.151 & 0.413 & -0.153 & 0.416 & -0.155 \\

Linear-Hidden (Reduce) & 100-shot
& 0.480 & 0.224 & 0.468 & 0.202 & 0.459 & 0.197 & 0.469 & 0.220 & 0.446 & 0.189 \\

\bottomrule
\end{tabular}

\caption{User-averaged PIAA performance on PARA and LAPIS.
Each model column reports Spearman’s $\rho$ and $R^2$.
Best values per column are highlighted in bold.}
\label{tab:para_lapis_useravg_metrics}
\end{table*}

%% file: figures/layerwise_piaa.tex
\begin{figure}[t]
  \centering
  \setlength{\tabcolsep}{2pt} %
  \renewcommand{\arraystretch}{1.0}

  \begin{tabular}{cc}
    \includegraphics[width=0.48\columnwidth]{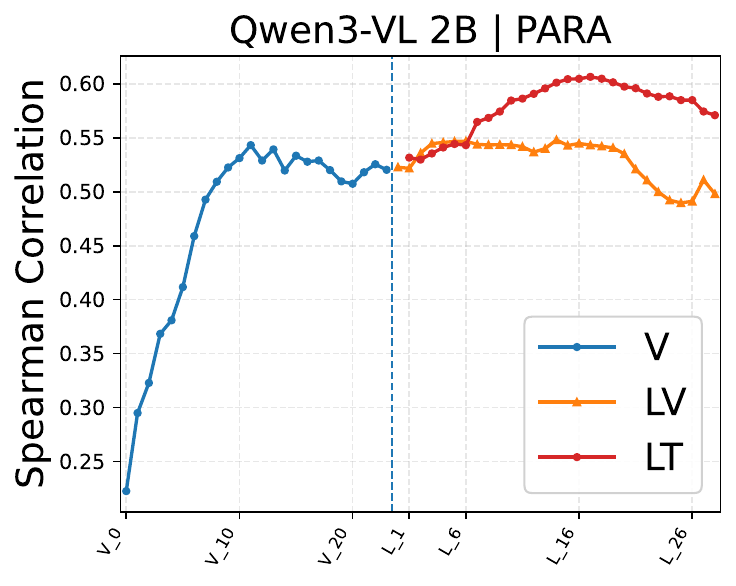} &
    \includegraphics[width=0.48\columnwidth]{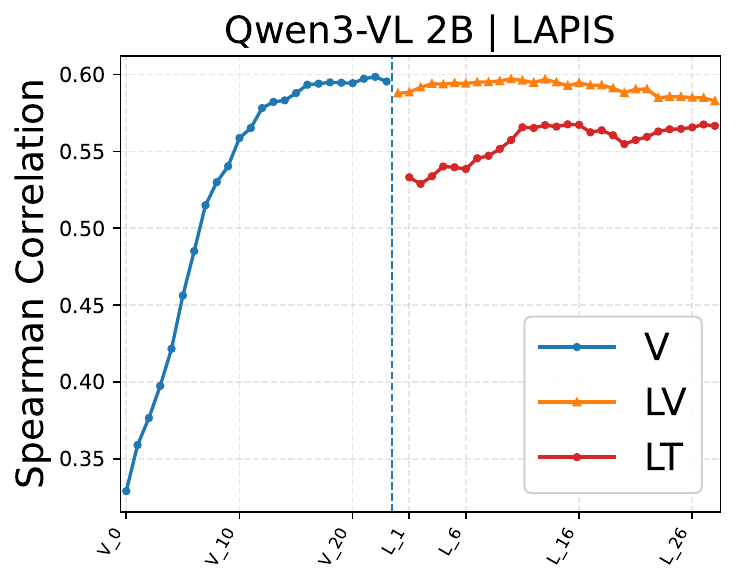} \\
    \includegraphics[width=0.48\columnwidth]{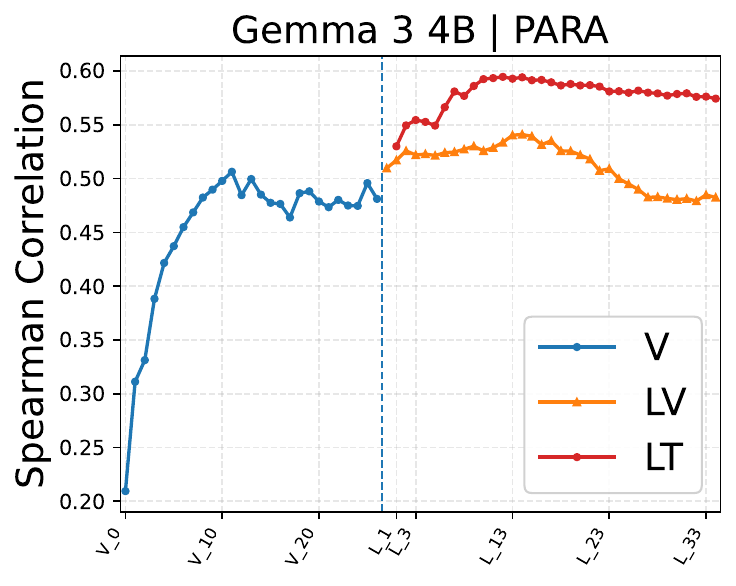} &
    \includegraphics[width=0.48\columnwidth]{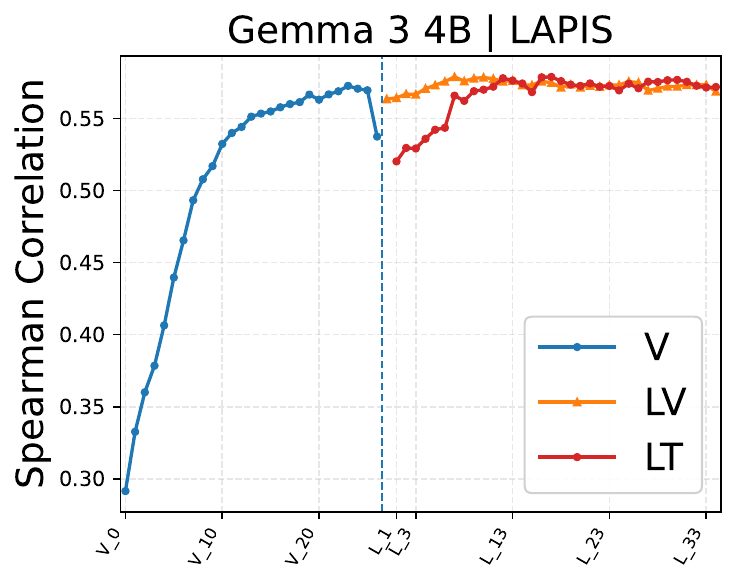}
  \end{tabular}

  \caption{Layer-wise PIAA performance across $\mathbf{V}$, $\mathbf{LV}$, and $\mathbf{LT}$ representations for multiple models and datasets.}
  \label{fig:layerwise_piaa}
\end{figure}

%% file: figures/repr_user.tex
\begin{figure*}[t]
  \centering
  \begin{minipage}[t]{0.333\textwidth}
    \centering
    \includegraphics[width=\linewidth]{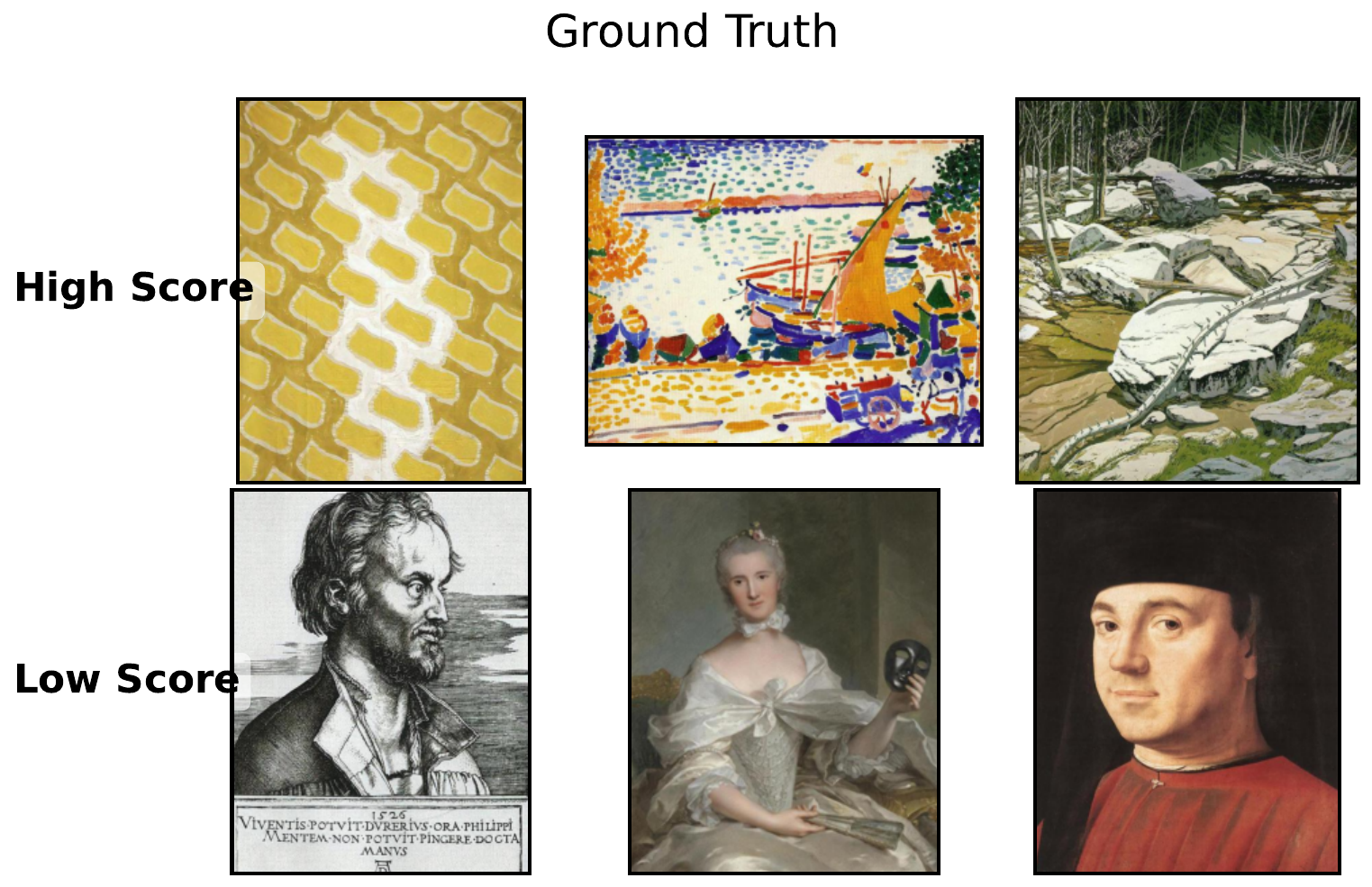}
  \end{minipage}%
  \begin{minipage}[t]{0.333\textwidth}
    \centering
    \includegraphics[width=\linewidth]{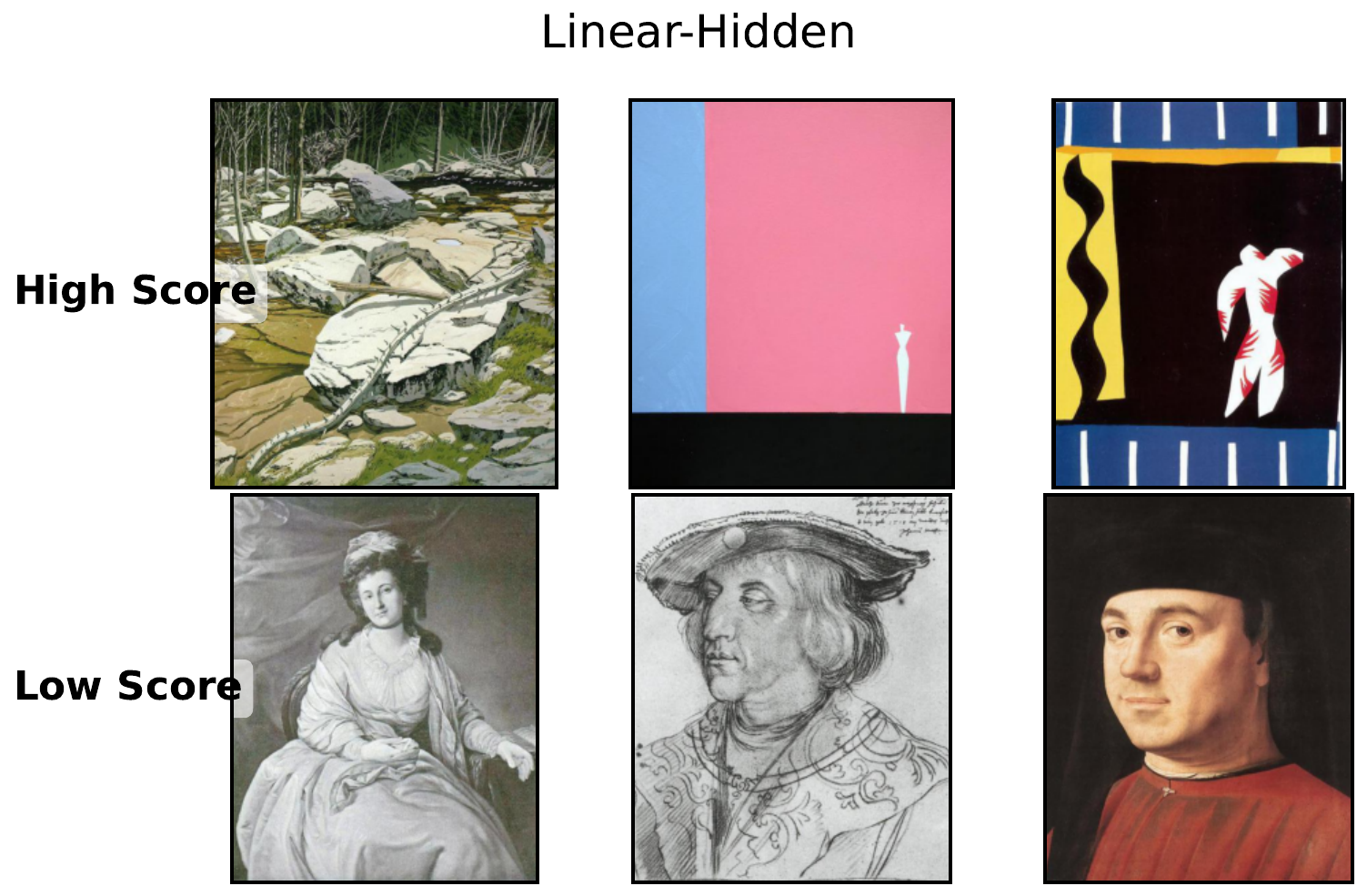}
  \end{minipage}%
  \begin{minipage}[t]{0.333\textwidth}
    \centering
    \includegraphics[width=\linewidth]{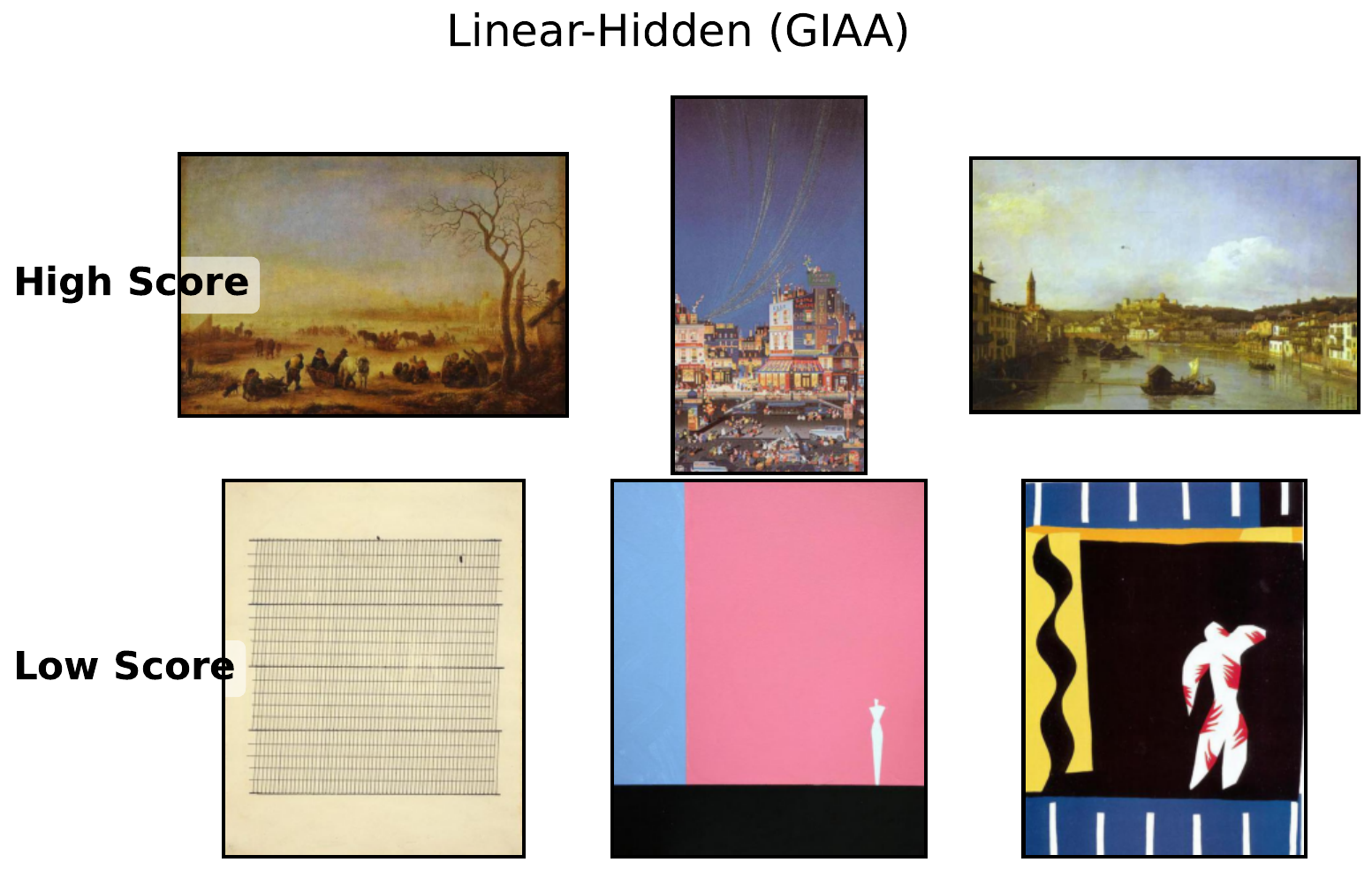}
  \end{minipage}

  \caption{Examples of the LAPIS images assigned high (top) and low (bottom) scores by different methods for a representative user.}
  \label{fig:repr_user}
\end{figure*}

%% file: tables/piaa_ici_in_domain.tex
\begin{table}[t]
\centering
\small
\setlength{\tabcolsep}{4pt}
\renewcommand{\arraystretch}{1.1}
\begin{tabular}{@{}l c@{}}
\toprule
Method & \multicolumn{1}{c}{PARA} \\
\midrule
Linear-Hidden & \rhorcell{0.611}{0.362} \\
PIAA-ICI (w/o User Attr) & \rhorcell{\textbf{0.620}}{0.392} \\
PIAA-ICI (w/ User Attr) & \rhorcell{0.619}{\textbf{0.424}} \\
\bottomrule
\end{tabular}
\caption{In-domain comparison on PARA. Linear-Hidden uses Qwen3-VL 4B with $\mathbf{LT}_{15}$. Values are reported as $\rho / R^2$.}
\label{tab:piaa_ici_in_domain}
\end{table}

%% file: tables/piaa_ici_cross_domain.tex
\begin{table}[t]
\centering
\small
\setlength{\tabcolsep}{4pt}
\renewcommand{\arraystretch}{1.1}

\begin{tabular}{@{}l c c@{}}
\toprule
Method & \multicolumn{1}{c}{PARA} & \multicolumn{1}{c}{LAPIS} \\
\midrule
Linear-Hidden 
& \rhorcell{0.611}{0.362} 
& \rhorcell{\textbf{0.568}}{\textbf{0.319}} \\

PIAA-ICI (w/o User Attr) 
& \rhorcell{\textbf{0.620}}{\textbf{0.392}} 
& \rhorcell{0.206}{-0.062} \\
\bottomrule
\end{tabular}

\caption{Cross-domain comparison. PIAA-ICI is pretrained on PARA, while Linear-Hidden uses pretrained VLM representations without dataset-specific pretraining. Values are reported as $\rho / R^2$.}
\label{tab:piaa_ici_cross_domain}
\end{table}

%% file: tables/hard_latex_table.tex
\begin{table*}[t]
\centering
\small
\setlength{\tabcolsep}{3pt}
\renewcommand{\arraystretch}{1.0}

\begin{tabular}{
@{}l l
*{5}{S[table-format=1.3] @{/} S[table-format=-1.3]}
@{}}
\toprule
Method & Support
& \multicolumn{6}{c}{Qwen3-VL}
& \multicolumn{4}{c}{Gemma 3} \\

& 
& \multicolumn{2}{c}{2B}
& \multicolumn{2}{c}{4B}
& \multicolumn{2}{c}{8B}
& \multicolumn{2}{c}{4B}
& \multicolumn{2}{c}{12B} \\

&
& \multicolumn{2}{c}{$\rho$ / $R^2$}
& \multicolumn{2}{c}{$\rho$ / $R^2$}
& \multicolumn{2}{c}{$\rho$ / $R^2$}
& \multicolumn{2}{c}{$\rho$ / $R^2$}
& \multicolumn{2}{c}{$\rho$ / $R^2$} \\
\midrule

Raw Text &
& 0.380 & -2.292
& 0.405 & -4.308
& 0.387 & -2.943
& 0.347 & -3.109
& 0.330 & -6.295 \\

\midrule

Linear-Hidden & 100-shot
& \bfseries 0.467 & 0.058
& \bfseries 0.472 & 0.053
& \bfseries 0.463 & 0.059
& \bfseries 0.463 & 0.079
& \bfseries 0.458 & 0.018 \\

Linear-Hidden (GIAA) & 100-shot
& 0.428 & -0.908
& 0.435 & -0.867
& 0.440 & -0.894
& 0.422 & -1.093
& 0.432 & -0.930 \\

Linear-Hidden (Reduce) & 100-shot
& 0.431 & \bfseries 0.150
& 0.447 & \bfseries 0.157
& 0.447 & \bfseries 0.085
& 0.436 & \bfseries 0.145
& 0.450 & \bfseries 0.159 \\

\bottomrule
\end{tabular}

\caption{PIAA performance on PARA for users with low agreement to GIAA (``hard'' users).}
\label{tab:hard_dataset_useravg_metrics}
\end{table*}

%% file: figures/annotator_agreement.tex
\begin{figure}[t]
  \centering
  \begin{minipage}[t]{0.5\linewidth}
    \centering
    \includegraphics[width=\linewidth]{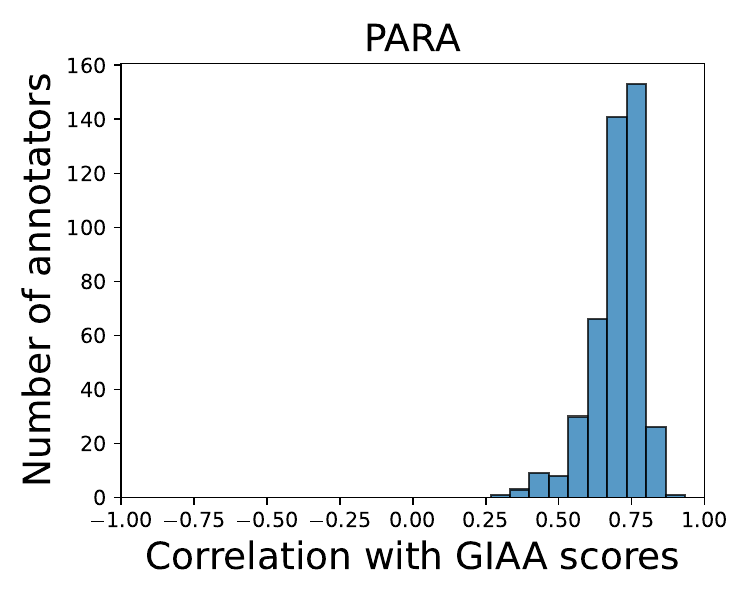}
  \end{minipage}%
  \begin{minipage}[t]{0.5\linewidth}
    \centering
    \includegraphics[width=\linewidth]{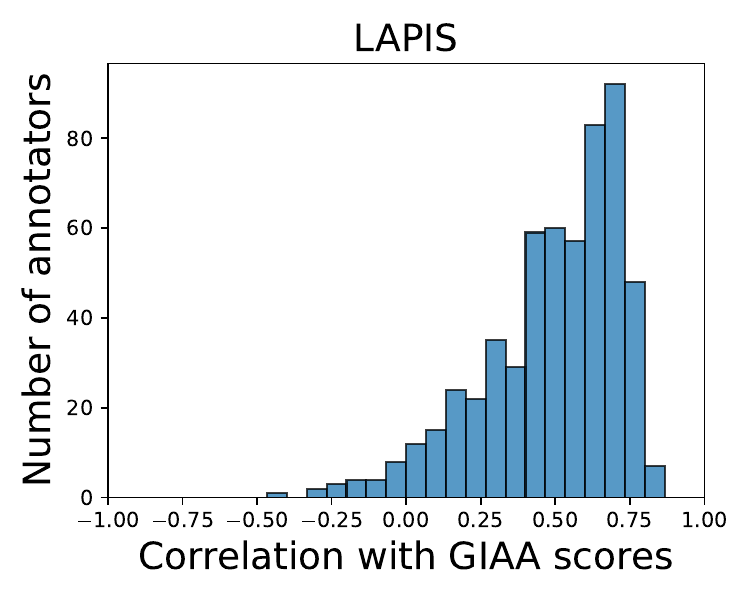}
  \end{minipage}%

  \caption{Distribution of Spearman correlation between PIAA and GIAA scores across users in PARA and LAPIS.}
  \label{fig:annotator_agreement}
\end{figure}

%% file: chapters/5_conclusion.tex
\section{Conclusion}
\label{sec:conclusion}
In this paper, we investigated what aesthetic attributes are encoded within VLMs and how such representations can be leveraged for PIAA.\par
Through linear probing, we demonstrated that VLMs encode a diverse set of aesthetic attributes, including those that propagate into language decoder layers.
We further demonstrated that these hidden representations provide effective signals for individual-level personalization.
Our analysis also revealed domain-dependent differences in how aesthetic attributes relevant to PIAA are represented across vision and language components of VLMs.\par
We identify two main directions for future work.
First, further investigation is needed to uncover additional aesthetic attributes that are not captured by current probes, particularly those encoded in vision tokens for artwork domains.
Second, an important next step is to translate these representation-level insights into improved personalization of text-based behaviors in VLMs, such as aesthetic judgments and caption generation.\par
We hope that our findings contribute to a deeper understanding of aesthetic representations in multimodal language models and the future development of VLMs that are better aligned with subjective, individual user preferences.

%% file: chapters/appendix.tex
\section{Implementation Details}
\label{appendix:detailed-experimental-settings}
\subsection{Models}
\label{appendix:models}
All experiments load VLMs using the Transformers~\cite{transformers} library.\footnote{\url{https://huggingface.co/collections/Qwen/qwen3-vl}}\footnote{\url{https://huggingface.co/docs/transformers/main/model_doc/dinov3}}\footnote{\url{https://huggingface.co/collections/google/gemma-3-release}}.
For experiments with high memory requirements, such as LoRA fine-tuning, we fix the floating-point precision to bfloat16.
For other experiments, we preserve the original precision of the released model weights by specifying ``auto'' for the \texttt{torch\_dtype} parameter.\par
For all experiments, we fix random seeds for data sampling and training procedures where applicable, and use deterministic decoding for text generation.

\subsection{Computational Resources}
\label{appendix:computational-resources}
All experiments were conducted on a single computing node equipped with one NVIDIA H100 GPU.
The few-shot (100-shot) baseline experiments reported in Appendix~\ref{appendix:full-piaa-results} required approximately 24 GPU-hours.
All other experiments, including linear probing and PIAA prediction, were completed within approximately 8 GPU-hours.

\subsection{Regressor Implementation}
\label{appendix:probing-implementation}

All experiments that involve ridge regression, including linear probing in Section~\ref{sec:probing} and the Linear-Hidden methods in Section~\ref{sec:piaa}, are implemented using a scikit-learn~\cite{sklearn} pipeline consisting of \textit{StandardScaler} followed by \textit{RidgeCV}.\par

Formally, the regression objective minimizes the following loss~\footnote{\url{https://scikit-learn.org/stable/modules/generated/sklearn.linear_model.Ridge.html}} with respect to the parameter vector $\mathbf{w}$:
\begin{align}
    \lVert \mathbf{y} - \mathbf{X}\mathbf{w} \rVert_2^2 + \alpha \lVert \mathbf{w} \rVert_2^2 ,
\end{align}
where $\mathbf{X}$ denotes the matrix of VLM features standardized to zero mean and unit variance, and $\mathbf{y}$ denotes the corresponding ground-truth labels.
The regularization coefficient $\alpha$ controls the strength of the L2 penalty.\par

In our experiments, $\alpha$ is selected via cross-validation on the training set from 13 logarithmically spaced candidates in the range $[10^{-3}, 10^{3}]$.

\subsection{PIAA Baselines}
\label{appendix:piaa-baseline-details}
\input{prompts/few-shot-prompt}
This section lists the baselines used in Section~\ref{sec:piaa} and provides implementation details for each method.

\paragraph{Raw Text}
For the \textbf{Raw Text} baseline, we prompt the VLM to output a general image aesthetics assessment (GIAA) score without any user-specific conditioning.
We use the following instruction to obtain a scalar score from the model:
\textit{``Assess the overall aesthetic quality of this image. Please rate it on a scale from 1 to 5. Output only the numeric score, and do not output any other text.''}

\paragraph{Adjust-Bias}
For the \textbf{Adjust-Bias} baseline, we first obtain GIAA predictions for both the support and test sets using the same prompt as in \textbf{Raw Text}.
We then estimate a user-specific bias term based on the support set and subtract it from test-time predictions to obtain personalized scores.\par

Formally, let $N$ denote the size of the support set, $I_i$ the $i$-th image in the support set, $s_i$ the score assigned by the target user $u$, and $v(I_i)$ the GIAA score predicted by the VLM.
The bias term $b_u$ and the personalized prediction $s_u(I)$ for a test image $I$ are computed as:
\begin{align}
   b_u &= \frac{1}{N}\sum_{i=1}^N \bigl(v(I_i) - s_i\bigr), \\
   s_u(I) &= v(I) - b_u .
\end{align}

\paragraph{Few-shot}
For the \textbf{Few-shot} baseline, we construct prompts that interleave example images with corresponding personalized scores from the support set, enabling the model to infer user-specific scoring tendencies from a small number of demonstrations.
Figure~\ref{prompt:few-shot} illustrates a pseudo-prompt showing how few-shot image-score pairs are integrated into the prompt.

\paragraph{LoRA}
For the \textbf{LoRA} baseline, we perform user-specific LoRA fine-tuning using the PEFT~\cite{peft} library.
We apply LoRA to all linear layers of the VLM.
Unless otherwise specified, we set the LoRA hyperparameters to $\alpha = 16$, rank $r = 8$, and dropout $= 0.1$, following standard practice.
All other parameters use the default settings provided by the PEFT library.\par

We use the AdamW optimizer~\cite{adamw} with a learning rate of $1 \times 10^{-4}$.
A linear learning-rate scheduler with warmup is applied using \texttt{get\_linear\_schedule\_with\_warmup} from Transformers~\cite{transformers}, with the warmup period set to 10\% of the total training steps.\par

For each user, we train the model for three epochs.
The batch size is set to 4 by default, but reduced to 2 for Qwen3-VL~8B due to memory constraints.\par

Due to the high resolution of images in PARA and the variable vision token length in Qwen3-VL, we resize images for the LoRA baseline on Qwen3-VL models such that the longer side does not exceed 1024 pixels.

\subsection{Aesthetic Attribute Distribution}
\label{appendix:dataset-stats}

\input{figures/aadb_attributes_hist}
\input{figures/para_attributes_hist}

Figures~\ref{fig:aadb-attributes} and \ref{fig:para-attributes} show the distributions of aesthetic attribute annotations in AADB and PARA, respectively.
In AADB, although annotations are provided as continuous values, several attributes such as \textit{MotionBlur}, \textit{Repetition}, and \textit{Symmetry} take on only a limited number of distinct values.
For these attributes, the most frequent value accounts for more than half of the annotations.
When interpreting the probing results, this characteristic should be taken into account, and direct comparisons across attributes based solely on correlation values should be avoided.\par

\input{figures/attr_correlation}

Figure~\ref{fig:attr_correlation} illustrates the Spearman correlation between aesthetic attributes in the two datasets.
While inter-attribute correlations are generally moderate in AADB, the aesthetic attributes annotated in PARA exhibit substantially higher correlations with each other.
We hypothesize that this difference stems from the annotation protocol used in PARA.
First, each attribute score is obtained by averaging subjective ratings from more than 20 annotators per image.
Second, all attribute annotations are collected within a single annotation interface.
These factors may encourage consistent scoring patterns across attributes, leading to higher inter-attribute correlations.\par

Due to this strong correlation structure, the general aesthetic attributes in PARA are less suitable for analyzing the diversity of aesthetic attributes encoded in VLMs, which is the primary focus of our probing experiments.
Accordingly, we adopt AADB as the primary dataset for the probing results reported in Section~\ref{section:probing-result}, and treat PARA-based probing results as supplementary analyses.

\subsection{Other Software and Artifacts}
\label{appendix:other-software}

All experiments were implemented in Python~3.12.11.
For VLM inference and training, we used PyTorch~\cite{pytorch}~2.6.0+cu126 and the Transformers~\cite{transformers} library version~4.57.1.
Additional libraries used in our experiments include Albumentations~\cite{albumentations}~2.0.8, \texttt{peft}~\cite{peft}~0.18.0, Pillow~\cite{pillow}~12.0.0, and OpenCV~\cite{opencv}~4.11.0.86.
Evaluation metrics were computed using scikit-learn~\cite{sklearn}~1.7.2 and SciPy~\cite{scipy}~1.16.3.

\section{Detailed Results}
\subsection{Full AADB Probing Results}
\label{appendix:full-aadb-probing}

\input{tables/probing_table_v}
\input{tables/probing_table_lv}
\input{tables/probing_table_ltau}

Tables~\ref{tab:probing_best_layers_v} and \ref{tab:probing_best_layers_lv} summarize the results of our probing experiments on AADB using $\mathbf{V}$ and $\mathbf{LV}$ representations, respectively.
While some attribute-specific differences can be observed (e.g., \textit{VividColor} is detected more strongly with $\mathbf{V}$ than with $\mathbf{LT}$), the overall trends are consistent across all three representation types.\par

We additionally examine $\mathbf{L\tau}_i$, defined as the representation of the last text token at the $i$-th language decoder layer.
The probing results for $\mathbf{L\tau}_i$ are shown in Table~\ref{tab:probing_best_layers_ltau}.
Consistent with the results for $\mathbf{V}$, $\mathbf{LV}$, and $\mathbf{LT}$, no substantial differences are observed in the overall probing trends.

\subsection{Full PIAA Experiment Results}
\label{appendix:full-piaa-results}
\input{tables/para_piaa_table}
\input{tables/lapis_piaa_table}

Tables~\ref{tab:para_useravg_metrics} and \ref{tab:lapis_useravg_metrics} present the complete results of our PIAA experiments, including both 10-shot and 100-shot settings.\par

Due to the considerable context length required for the \textbf{Few-shot} baseline in the 100-shot setting, we initially evaluated this configuration using the relatively lightweight Gemma~3~4B model.
After confirming that the 100-shot Few-shot setting did not yield substantial performance improvements over the 10-shot setting, we did not run additional 100-shot Few-shot experiments for other models.\par

When comparing the 10-shot and 100-shot results, all Linear-Hidden variants consistently benefit from larger support sets.
However, for the PARA dataset, the 10-shot Linear-Hidden results are lower than those of the \textbf{Raw Text} baseline.
We attribute this behavior to the fact that VLMs capture GIAA aspects of photographs more effectively than those of artworks, and that annotator agreement with GIAA scores in PARA is relatively strong.

\highlight
\subsection{PIAA Results on Vision Encoder Features}
\label{appendix:vision-encoder-feature-piaa}

\input{tables/piaa-vision-encoder}

We also report PIAA performance under linear personalization using vision encoder features, inspired by prior work~\cite{clip-knows-image-aesthetics}, which performs GIAA based on CLIP~\cite{clip} representations.

Both Qwen3-VL and Gemma~3 employ vision encoders derived from the SigLIP family~\cite{siglip}, which is closely related to CLIP-style contrastive pretraining.
We therefore use the final vision encoder layers as CLIP-like features: specifically, $V_{23}$ for Qwen3-VL 2B and $V_{26}$ for Gemma~3 4B.

These representations are used in the same linear personalization framework as Linear-Hidden and Linear-Hidden (Reduce) described in Section~\ref{sec:piaa}, allowing us to evaluate the effect of projecting into the 11-dimensional aesthetic subspace.
All evaluations are conducted under the 100-shot setting on both PARA and LAPIS.

The results are shown in Table~\ref{tab:piaa_vision_encoder}.
On PARA, reducing the representation to the 11-dimensional aesthetic subspace degrades performance compared to using the full vision features.
In contrast, when using the language-decoder representation ($\mathbf{LT}_{15}$, Table~\ref{tab:para_lapis_useravg_metrics}), no comparable degradation is observed.

One possible explanation is that the final vision encoder layers provide weaker representations of certain aesthetic attributes.
As shown in Figure~\ref{fig:probing}, attributes such as Light and Repetition are better captured in intermediate vision encoder layers or in language decoder layers than in the final vision encoder layer.
The under-representation of these attributes may contribute to the observed performance degradation.

\subsection{Validation of Undefined Correlation Values}
\label{appendix:validate-undefined-correlations}
\input{tables/combined_rho_nan_counts}

As mentioned in Section~\ref{sec:piaa-other-settings}, we exclude users for whom the correlation between model predictions and ground-truth scores is undefined when computing user-averaged PIAA performance (Tables~\ref{tab:para_useravg_metrics} and \ref{tab:lapis_useravg_metrics}).
To validate this design choice, we analyze how frequently undefined correlation values occur across methods. \par

The results are summarized in Table~\ref{tab:combined_rho_nan_count}.
As shown, a substantial number of undefined correlations are observed only for the \textbf{Few-shot} baseline.
This issue is not observed in the Raw Text baseline, suggesting that few-shot prompting for PIAA often causes VLMs to produce nearly constant predictions, which leads to undefined correlation values. \par

The Linear-Hidden variants under the 10-shot setting on PARA also exhibit a small number of undefined cases.
We verified that, for these users, the support sets contain nearly constant ground-truth labels, causing the learned regression function to collapse to a constant predictor. \par

Overall, these observations do not affect our main conclusion regarding the superiority of the Linear-Hidden method, as the number of affected users is minimal and can be attributed to label distribution rather than model behavior.

\subsection{Statistical Validation of the Results}
\label{sec:statistical-validation}

\input{tables/para_rho_ci}

\input{tables/lapis_rho_ci}

We assess the statistical robustness of our main PIAA results (Table~\ref{tab:para_lapis_useravg_metrics}) using bootstrap resampling.
Specifically, we perform 2,000 bootstrap resamples of the 200 test users with replacement.
For each resampled set, we compute the mean per-user Spearman correlation ($\rho$), and estimate 95\% confidence intervals (CI) from the empirical bootstrap distribution.

The results for PARA and LAPIS are presented in Tables~\ref{tab:para_rho_ci} and \ref{tab:lapis_rho_ci}.
Across all models, the confidence intervals remain narrow and are consistent with the reported mean values in Table~\ref{tab:para_lapis_useravg_metrics}, indicating that the results are stable under resampling.

We further conduct pairwise comparisons between Linear-Hidden and all text-based baselines (Raw Text, Few-shot, Adjust-Bias, LoRA), as well as their variants (GIAA, Reduce).
For each bootstrap resample, we compute the difference
\[
\Delta = \rho_{\text{baseline}} - \rho_{\text{Linear-Hidden}},
\]
and estimate the empirical probability $P(\Delta > 0)$.

We observe:
\begin{itemize}
    \item For PARA: $P(\Delta > 0) = 0$ for all text-based baselines.
    \item For LAPIS: $P(\Delta > 0) = 0$ for all baselines, including Linear-Hidden variants.
\end{itemize}

These results indicate that Linear-Hidden consistently outperforms competing methods across bootstrap resamples, providing strong statistical evidence for the performance improvements reported in the main paper.

\finishhighlight

\section{Additional Experiments}
\subsection{Probing with Augmented Images}
\label{appendix:probing-aug}

\input{figures/sample_aug}
\input{figures/probing_aug_attr}

Since the aesthetic attributes used for probing on AADB exhibit non-negligible correlations, the results presented in Section~\ref{section:probing-result} may be driven by a limited set of underlying visual traits.
To examine this possibility, we conduct an additional analysis based on controlled image augmentations.

Specifically, we apply the following two augmentations to images in AADB and repeat the same probing experiments:
\begin{itemize}
    \item \textbf{Grayscale}: Input images are converted to grayscale using Pillow~\cite{pillow}, largely removing color information while preserving overall image structure.
    \item \textbf{Thin Plate Spline}: Thin Plate Spline transformations are applied using Albumentations~\cite{albumentations} to introduce geometric distortions while approximately preserving color statistics.
\end{itemize}
\par

Figure~\ref{fig:probing-aug-attr} shows probing results of Qwen3-VL~2B for the \textit{Object} and \textit{VividColor} attributes under different augmentations.
For the \textit{Object} attribute, performance degrades substantially under Thin Plate Spline augmentation but remains relatively stable under Grayscale conversion.
In contrast, probing performance for \textit{VividColor} drops significantly under Grayscale while being less affected by Thin Plate Spline.\par

These contrasting behaviors indicate that different aesthetic attributes rely on distinct visual traits and are differentially affected by color and geometric transformations.
This observation supports our claim that the probing results in Section~\ref{section:probing-result} reflect the presence of multiple, disentangled aesthetic attributes encoded in VLM representations, rather than being driven by a single correlated feature.

\subsection{Probing on PARA}
\label{appendix:probing-on-para}

\input{tables/probing_table_para}
\input{figures/probing-para}

As described in Section~\ref{sec:probing-settings}, we also perform linear probing on the general aesthetic attributes provided in the PARA dataset.
The results are summarized in Table~\ref{tab:probing_best_layers_para} and visualized in Figure~\ref{fig:probing-para}.\par

Consistent with our findings on AADB, the models strongly encode PARA aesthetic attributes, including in the language decoder layers.
However, as discussed in Appendix~\ref{appendix:dataset-stats}, the aesthetic attributes in PARA exhibit strong inter-attribute correlations.
As a result, these probing results alone do not provide strong evidence for the presence of diverse and disentangled aesthetic attributes encoded in VLM representations.

\subsection{Prompt Sensitivity of the Probing Results}
\label{appendix:prompt-sensitivity}

\input{tables/probing_prompts_table}

While Section~\ref{sec:probing} demonstrates that VLMs encode diverse aesthetic attributes, it remains unclear to what extent these representations are sensitive to the specific instructions provided to the model.
Such sensitivity could potentially affect the validity of the probing results, as well as downstream experiments in Section~\ref{sec:piaa}, where different prompts are required for baselines such as \textbf{Few-shot}.\par

To examine this issue, we repeat the probing experiments from Section~\ref{sec:probing} on AADB using Qwen3-VL~2B with several different instruction variants:

\paragraph{Base}
The same instruction used in Section~\ref{sec:probing}:
\textit{``Assess the aesthetics of this image.''}

\paragraph{Numeric}
An instruction that explicitly enforces numeric output formatting:
\textit{``
Assess the aesthetics of this image.
Please rate it on a scale from 1 to 5.
Output only the numeric score, and do not output any other text.
''}

\paragraph{Attribute}
An instruction that explicitly lists aesthetic attribute names to encourage attribute-aware assessment:
\textit{``
Assess the aesthetics of this image with respect to the following attributes:
\{attrs\}.
You do not need to output the attributes explicitly; use them only as internal criteria.
''}
Here, \textit{\{attrs\}} denotes a concatenation of attribute names (e.g., \textit{BalancingElements, ColorHarmony, \ldots}).

\paragraph{Unrelated}
An instruction that is unrelated to aesthetic assessment:
\textit{``Describe the weather today in one sentence.''}

The results are summarized in Table~\ref{tab:probing_best_by_prompt}.
Across all prompt variants, the final probing performance exhibits no significant differences.
Based on this observation, we conclude that the probing results are robust to reasonable variations in prompt design, and we therefore adopt flexible prompt formulations in Section~\ref{sec:piaa}.

\subsection{Effect of Image Resizing on PIAA}
\label{appendix:image-resizing-effect}
\input{tables/para_resize_table}
As described in Appendix~\ref{appendix:piaa-baseline-details}, we resize PARA images when running the \textbf{LoRA} baseline with Qwen3-VL models due to memory constraints.
To validate that this design choice does not significantly affect the PIAA results, we additionally evaluate the \textbf{Raw Text} baseline using the resized images and compare its performance with that obtained on the original images.\par

Table~\ref{tab:para_resize_table} summarizes the results.
Across all model sizes, the \textbf{Raw Text} baseline shows no substantial differences in either Spearman correlation or $R^2$ between the original and resized images.
This observation indicates that image resizing alone does not materially affect PIAA performance for Qwen3-VL models on PARA, thereby supporting the validity of the resizing strategy adopted for the \textbf{LoRA} baseline in Section~\ref{sec:piaa}.

\highlight
\subsection{PIAA with Combined Representations}
\label{sec:piaa-pair}
\input{figures/combined_heatmaps}

Given that Figure~\ref{fig:probing} suggests that the vision encoder and the language decoder capture complementary image representations, we investigate whether combining hidden representations from different components of VLMs can improve the PIAA performance of Linear-Hidden.

More specifically, we consider combinations of $\mathbf{V}_{5i}$ and $\mathbf{LT}_{5j}$ (for $i, j \in \mathbb{N}$), extracted from Gemma 3 4B and Qwen3-VL 2B, respectively, and evaluate Linear-Hidden PIAA performance on the LAPIS dataset.

The user-averaged Spearman correlation values obtained from the combined representations are shown in Figure~\ref{fig:combined_heatmaps}.
The heatmaps demonstrate that incorporating deeper-layer $\mathbf{V}_i$ representations alongside $\mathbf{LT}_i$ representations consistently improves PIAA performance.
In particular, for both models, the combination $(\mathbf{LT}_{15}, \mathbf{V}_{20})$ outperforms the $\mathbf{LT}_{15}$-only result reported in Table~\ref{tab:para_lapis_useravg_metrics}.

These results suggest that concatenating complementary representations from different components of VLMs can further enhance personalization performance.

\subsection{Domain Transferability of Linear-Hidden Feature Extraction}
\label{sec:piaa-domain-transfer}
\input{tables/logo-holdout-summary}

We conduct an experiment to evaluate the generalizability of the PIAA-related features learned by Linear-Hidden.

To the best of our knowledge, no PIAA dataset provides consistent annotator identities across distinct image domains such as photographs and artworks.
To approximate such a cross-domain setting, we use the \textit{2\_styles} attribute in the LAPIS dataset to partition images into two groups: ABSTRACT and FIGURATIVE.

Specifically, we select the same set of 200 users as in Section~\ref{sec:piaa}, combine their support set and test set to obtain 150 images per user, and split these images according to the \textit{2\_styles} attribute.
We then train Linear-Hidden on one style and evaluate it on the other, simulating a cross-domain transfer scenario.

The results are shown in Table~\ref{tab:logo-holdout-summary}.
We observe a substantial performance drop under this setting, with performance only marginally better than that of text-based baselines.
This suggests that the learned PIAA representations are not fully transferable across distinct image styles.

We propose two possible explanations for this phenomenon:
\begin{itemize}
    \item Information about a user's preferred image style plays a critical role in PIAA prediction.
    \item The relationship between low-level image attributes and user preferences differs across image styles.
\end{itemize}

We leave a more detailed investigation of these hypotheses for future work.

\finishhighlight

\section{License and Intended Use of Scientific Artifacts}
All scientific artifacts used in this work, including datasets, pretrained models, and software libraries, are utilized in accordance with their respective licenses and terms of use.
This study does not release new datasets or models.
The experiments are conducted solely for academic research purposes, and no artifacts are used in a manner that violates their original licensing conditions.

\section{AI Assistance Usage}
AI-assisted tools, including ChatGPT\footnote{\url{https://chatgpt.com/}} and Google Gemini\footnote{\url{https://gemini.google.com/}}, were used to support writing refinement and code development in accordance with ACL Policy on AI Writing/Coding Assistance.

%% file: prompts/few-shot-prompt.tex
\begin{figure}
\begin{Prompt}
\begin{verbatim}
You are an expert judge of image aesthetics.
I will show you some example images with
this user's ratings on a 1 to 5 scale.
From these examples, infer the user's
personal  preferences.
Then I will show you a new image; please
predict this user's rating for it.
For the examples, each rating is this user's
own rating, already mapped to a 1 to 5 scale.
When you answer for the final image, respond
with a single number from 1 to 5, and
nothing else.

{%
{{ images[idx] }}
Example {{ idx + 1}}. This user rated this
image {{ images[idx].score }} out of 5.
{%

Now, based on the user's previous ratings,
what is this user's rating for THIS image?
Answer with a single number from 1 to 5,
and do not output any other text.
{{ target_image }}
\end{verbatim}
\end{Prompt}
\caption{Prompt template used for the Few-shot PIAA baseline.}
\label{prompt:few-shot}
\end{figure}

%% file: figures/aadb_attributes_hist.tex
\begin{figure}[t]
  \centering
    \includegraphics[width=\linewidth]{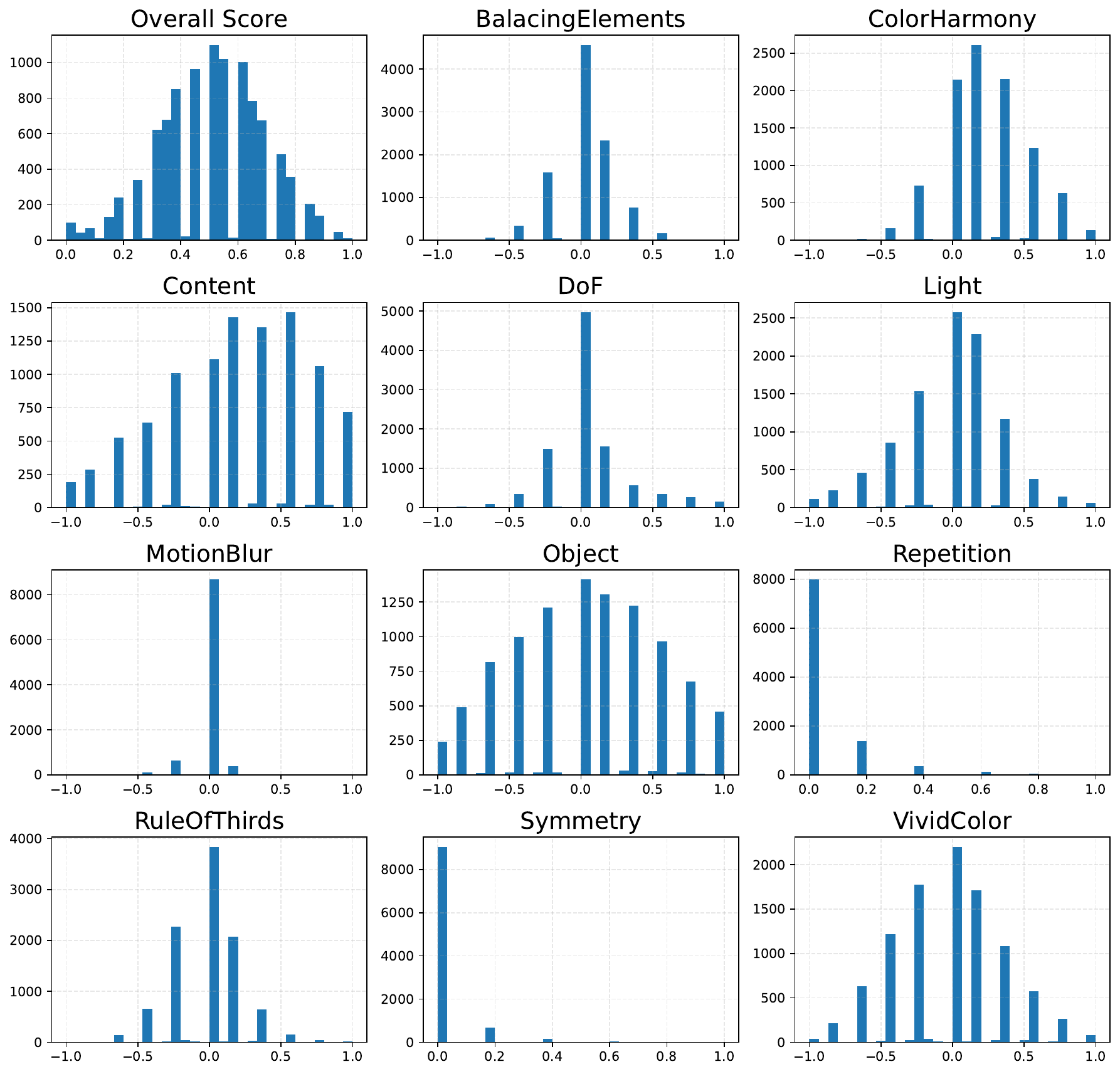}
  \caption{Distribution of aesthetic attribute annotations in AADB.}
  \label{fig:aadb-attributes}
\end{figure}

%% file: figures/para_attributes_hist.tex
\begin{figure}[t]
  \centering
    \includegraphics[width=\linewidth]{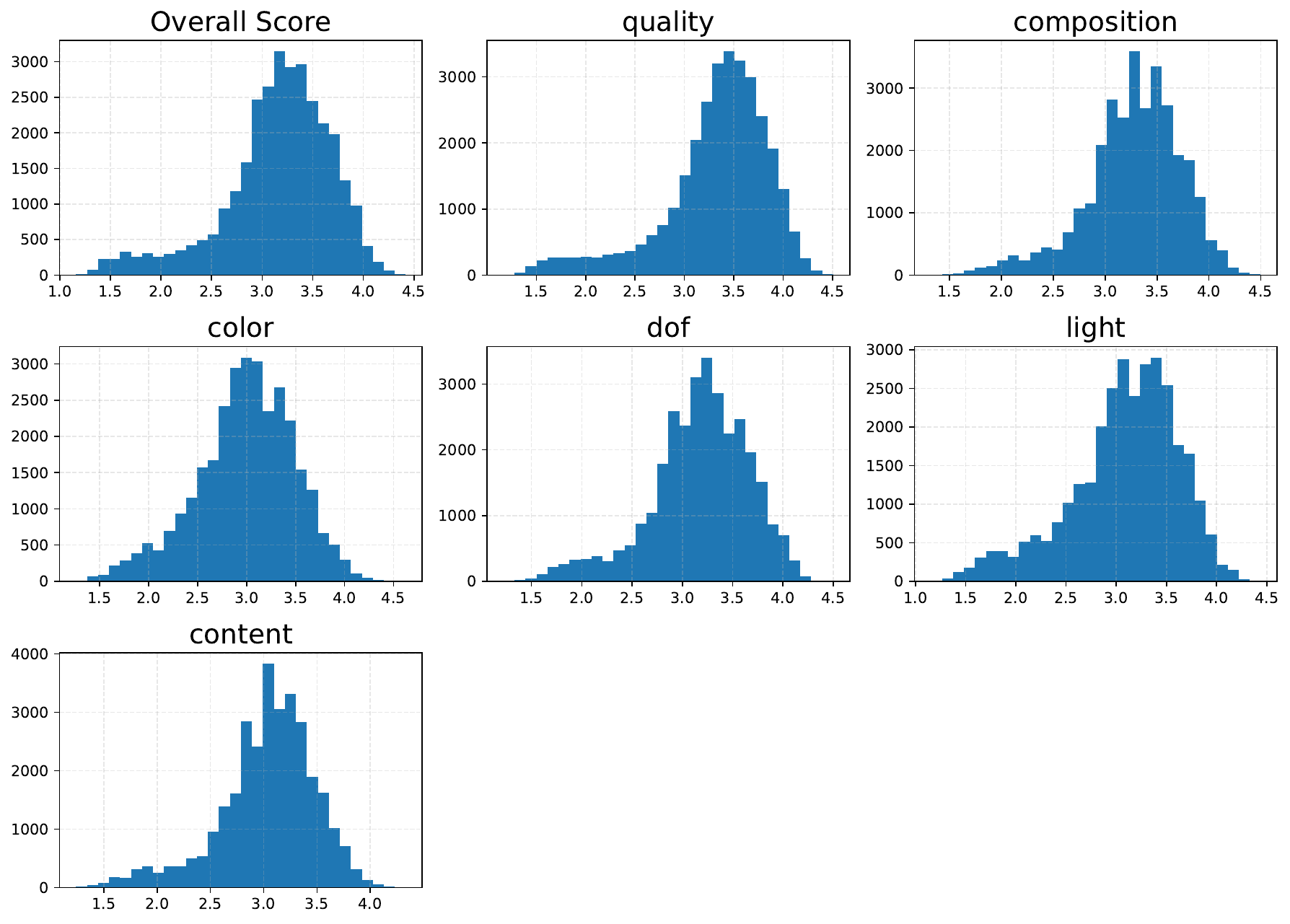}
  \caption{Distribution of aesthetic attribute annotations in PARA.}
  \label{fig:para-attributes}
\end{figure}

%% file: figures/attr_correlation.tex
\begin{figure}[t]
  \centering
    \includegraphics[width=\linewidth]{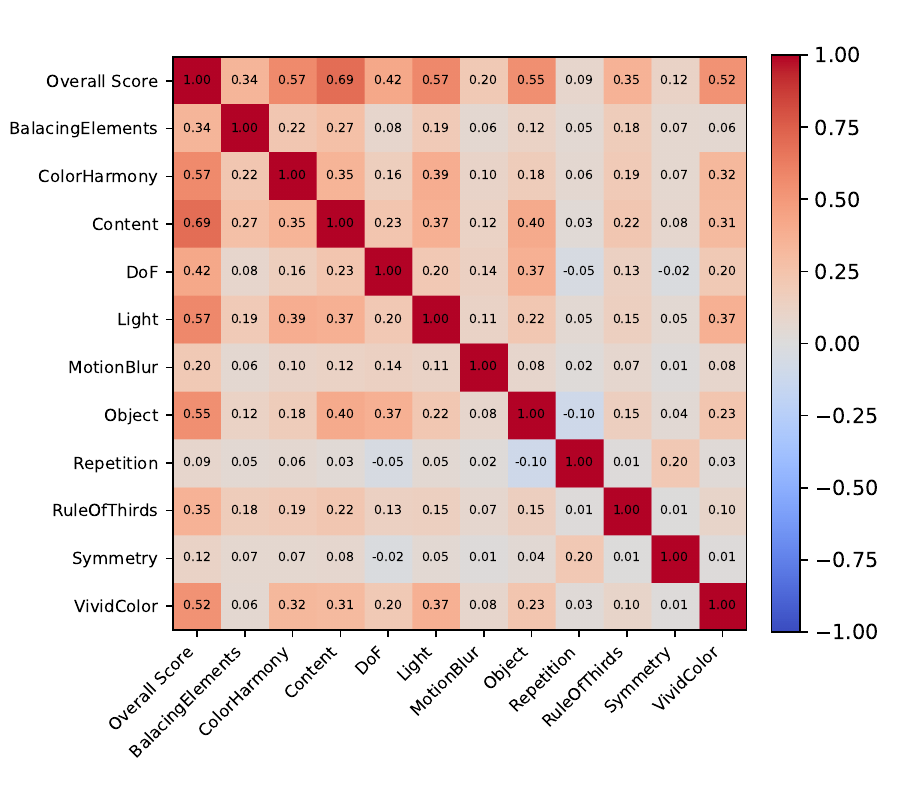}
    \includegraphics[width=\linewidth]{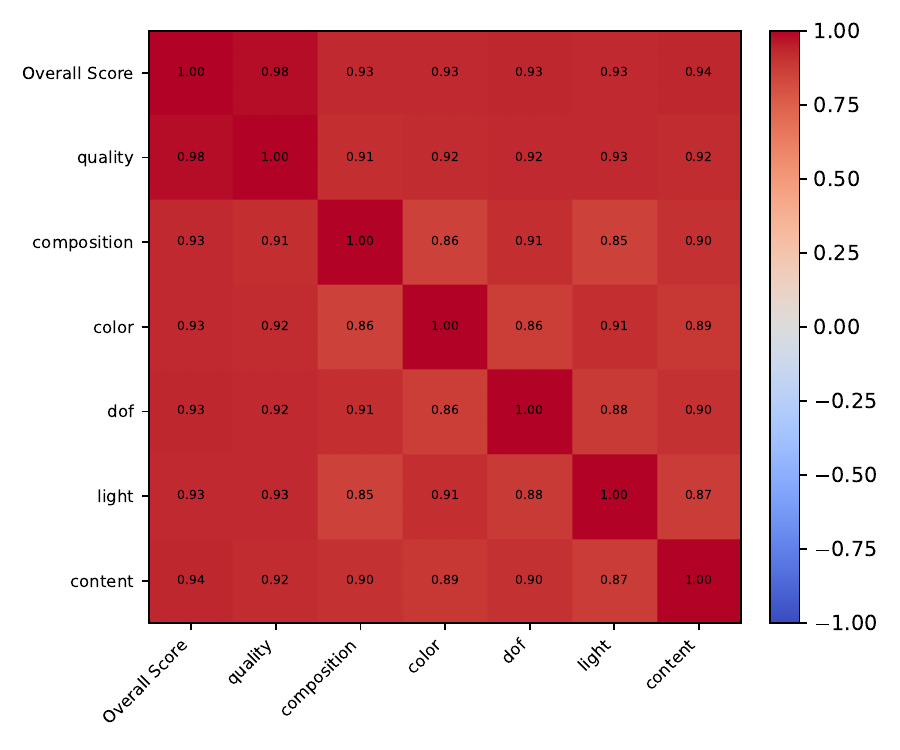}

  \caption{Spearman correlation between aesthetic attributes in AADB (top) and PARA (bottom).}
  \label{fig:attr_correlation}
\end{figure}

%% file: tables/probing_table_v.tex
\begin{table*}[t]
\centering
\small
\setlength{\tabcolsep}{3pt}
\renewcommand{\arraystretch}{1.0}
\begin{tabular}{
l
S @{\,(} c @{)\hspace{1em}}
S @{\,(} c @{)\hspace{1em}}
S @{\,(} c @{)\hspace{1em}}
S @{\,(} c @{)\hspace{1em}}
S @{\,(} c @{)\hspace{1em}}
S @{\,(} c @{)\hspace{1em}}
S @{\,(} c @{)}
}
\toprule
Attribute
& \multicolumn{6}{c}{Qwen3-VL}
& \multicolumn{4}{c}{Gemma 3}
& \multicolumn{4}{c}{DINOv3} \\

& \multicolumn{2}{c}{2B}
& \multicolumn{2}{c}{4B}
& \multicolumn{2}{c}{8B}
& \multicolumn{2}{c}{4B}
& \multicolumn{2}{c}{12B}
& \multicolumn{2}{c}{ViT-B/16}
& \multicolumn{2}{c}{ViT-L/16} \\
\midrule
BalancingElements & 0.325 & 21 & \bfseries 0.328 & 23 & 0.307 & 26 & 0.300 & 26 & 0.300 & 26 & 0.317 & 11 & 0.307 & 22 \\
ColorHarmony      & \bfseries 0.516 & 23 & 0.513 & 23 & 0.508 & 22 & 0.502 & 10 & 0.502 & 10 & 0.479 & 6  & 0.482 & 10 \\
Content           & \bfseries 0.621 & 23 & 0.621 & 23 & 0.608 & 26 & 0.579 & 26 & 0.579 & 26 & 0.551 & 12 & 0.579 & 17 \\
DoF               & 0.532 & 9  & 0.530 & 9  & \bfseries 0.535 & 12 & 0.513 & 10 & 0.513 & 10 & 0.506 & 7  & 0.507 & 18 \\
Light             & 0.488 & 15 & \bfseries 0.500 & 12 & 0.497 & 14 & 0.474 & 10 & 0.474 & 10 & 0.439 & 8  & 0.436 & 11 \\
MotionBlur        & 0.161 & 17 & 0.152 & 7  & \bfseries 0.180 & 10 & 0.139 & 19 & 0.139 & 19 & 0.161 & 7  & 0.143 & 3 \\
Object            & 0.709 & 13 & \bfseries 0.709 & 23 & 0.705 & 15 & 0.685 & 15 & 0.685 & 15 & 0.688 & 12 & 0.696 & 19 \\
Repetition        & 0.458 & 12 & 0.461 & 12 & \bfseries 0.462 & 26 & 0.446 & 15 & 0.446 & 15 & 0.438 & 8  & 0.451 & 19 \\
RuleOfThirds      & \bfseries 0.282 & 14 & 0.272 & 11 & 0.278 & 17 & 0.247 & 13 & 0.247 & 13 & 0.230 & 9  & 0.230 & 20 \\
Symmetry          & \bfseries 0.319 & 23 & 0.304 & 12 & 0.279 & 26 & 0.301 & 9  & 0.301 & 9  & 0.299 & 5  & 0.313 & 11 \\
VividColor        & 0.717 & 12 & \bfseries 0.719 & 13 & 0.718 & 13 & 0.698 & 13 & 0.698 & 13 & 0.686 & 3  & 0.685 & 10 \\
Overall Score     & 0.706 & 23 & \bfseries 0.707 & 23 & 0.701 & 22 & 0.673 & 25 & 0.673 & 25 & 0.636 & 9  & 0.666 & 17 \\
\bottomrule
\end{tabular}

\caption{Highest Spearman correlation achieved by linear probing on $\mathbf{V}$ layers for each aesthetic attribute in AADB.
Values in parentheses indicate the corresponding layer indices.}
\label{tab:probing_best_layers_v}
\end{table*}

%% file: tables/probing_table_lv.tex
\begin{table*}[t]
\centering
\small
\setlength{\tabcolsep}{3pt}
\renewcommand{\arraystretch}{1.0}

\begin{tabular}{
l
S @{\,(} c @{)\hspace{1em}}
S @{\,(} c @{)\hspace{1em}}
S @{\,(} c @{)\hspace{1em}}
S @{\,(} c @{)\hspace{1em}}
S @{\,(} c @{)}
}
\toprule

Attribute
& \multicolumn{6}{c}{Qwen3-VL}
& \multicolumn{4}{c}{Gemma 3} \\

& \multicolumn{2}{c}{2B}
& \multicolumn{2}{c}{4B}
& \multicolumn{2}{c}{8B}
& \multicolumn{2}{c}{4B}
& \multicolumn{2}{c}{12B} \\

\midrule

BalancingElements & 0.331 & 7  & \bfseries 0.349 & 13 & 0.314 & 23 & 0.307 & 1  & 0.300 & 1 \\
ColorHarmony      & 0.517 & 9  & \bfseries 0.519 & 1  & 0.517 & 9  & 0.478 & 21 & 0.489 & 12 \\
Content           & \bfseries 0.635 & 9  & 0.627 & 1  & 0.622 & 26 & 0.600 & 18 & 0.599 & 19 \\
DoF               & 0.533 & 6  & \bfseries 0.534 & 3  & 0.516 & 1  & 0.471 & 14 & 0.478 & 17 \\
Light             & 0.470 & 10 & \bfseries 0.478 & 11 & 0.474 & 8  & 0.423 & 20 & 0.428 & 7 \\
MotionBlur        & \bfseries 0.158 & 18 & 0.147 & 25 & 0.132 & 5  & 0.118 & 15 & 0.137 & 27 \\
Object            & \bfseries 0.716 & 14 & 0.712 & 2  & 0.710 & 2  & 0.695 & 1  & 0.692 & 21 \\
Repetition        & 0.442 & 2  & 0.446 & 3  & \bfseries 0.446 & 7  & 0.400 & 2  & 0.405 & 15 \\
RuleOfThirds      & 0.301 & 14 & \bfseries 0.303 & 25 & 0.285 & 2  & 0.257 & 13 & 0.247 & 16 \\
Symmetry          & \bfseries 0.320 & 3  & 0.304 & 16 & 0.287 & 16 & 0.280 & 25 & 0.291 & 26 \\
VividColor        & 0.710 & 2  & 0.714 & 3  & \bfseries 0.718 & 8  & 0.668 & 4  & 0.675 & 2 \\
Overall Score     & \bfseries 0.729 & 8  & 0.721 & 11 & 0.718 & 1  & 0.685 & 2  & 0.694 & 2 \\

\bottomrule
\end{tabular}

\caption{Highest Spearman correlation achieved by linear probing on $\mathbf{LV}$ layers for each aesthetic attribute in AADB.
Values in parentheses indicate the corresponding layer indices.}
\label{tab:probing_best_layers_lv}
\end{table*}

%% file: tables/probing_table_ltau.tex
\begin{table*}[t]
\centering
\small
\setlength{\tabcolsep}{3pt}
\renewcommand{\arraystretch}{1.0}

\begin{tabular}{
l
S @{\,(} c @{)\hspace{1em}}
S @{\,(} c @{)\hspace{1em}}
S @{\,(} c @{)\hspace{1em}}
S @{\,(} c @{)\hspace{1em}}
S @{\,(} c @{)}
}
\toprule

Attribute
& \multicolumn{6}{c}{Qwen3-VL}
& \multicolumn{4}{c}{Gemma 3} \\

& \multicolumn{2}{c}{2B}
& \multicolumn{2}{c}{4B}
& \multicolumn{2}{c}{8B}
& \multicolumn{2}{c}{4B}
& \multicolumn{2}{c}{12B} \\

\midrule

BalancingElements & 0.303 & 5  & \bfseries 0.322 & 6  & 0.295 & 1  & 0.306 & 14 & 0.302 & 14 \\
ColorHarmony      & 0.511 & 11 & \bfseries 0.526 & 22 & 0.526 & 28 & 0.493 & 12 & 0.496 & 15 \\
Content           & \bfseries 0.629 & 11 & 0.625 & 13 & 0.618 & 25 & 0.612 & 6  & 0.621 & 34 \\
DoF               & \bfseries 0.531 & 15 & 0.520 & 13 & 0.521 & 34 & 0.500 & 1  & 0.524 & 15 \\
Light             & 0.483 & 16 & \bfseries 0.510 & 26 & 0.485 & 25 & 0.442 & 10 & 0.469 & 5 \\
MotionBlur        & 0.154 & 2  & 0.155 & 29 & 0.169 & 4  & \bfseries 0.174 & 21 & 0.161 & 43 \\
Object            & 0.711 & 16 & \bfseries 0.719 & 19 & 0.717 & 13 & 0.705 & 18 & 0.711 & 10 \\
Repetition        & 0.453 & 7  & \bfseries 0.456 & 18 & 0.456 & 13 & 0.422 & 16 & 0.421 & 29 \\
RuleOfThirds      & 0.277 & 13 & \bfseries 0.294 & 33 & 0.268 & 3  & 0.264 & 22 & 0.257 & 16 \\
Symmetry          & 0.309 & 5  & \bfseries 0.315 & 18 & 0.306 & 12 & 0.267 & 7  & 0.292 & 33 \\
VividColor        & 0.680 & 18 & 0.692 & 3  & \bfseries 0.697 & 21 & 0.670 & 7  & 0.685 & 18 \\
Overall Score     & 0.713 & 11 & \bfseries 0.726 & 24 & 0.717 & 23 & 0.693 & 3  & 0.714 & 15 \\

\bottomrule
\end{tabular}

\caption{Highest Spearman correlation achieved by linear probing on $\mathbf{L\tau}$ layers for each aesthetic attribute in AADB.
Values in parentheses indicate the corresponding layer indices.}
\label{tab:probing_best_layers_ltau}
\end{table*}

%% file: tables/para_piaa_table.tex
\begin{table*}[t]
\centering
\small
\setlength{\tabcolsep}{3pt}
\renewcommand{\arraystretch}{1.0}

\begin{tabular}{
@{}l l
*{5}{S[table-format=1.3] @{/} S[table-format=-1.3]}
@{}}
\toprule
Method & Support
& \multicolumn{6}{c}{Qwen3-VL}
& \multicolumn{4}{c}{Gemma 3} \\

&
& \multicolumn{2}{c}{2B}
& \multicolumn{2}{c}{4B}
& \multicolumn{2}{c}{8B}
& \multicolumn{2}{c}{4B}
& \multicolumn{2}{c}{12B} \\

&
& \multicolumn{2}{c}{$\rho$ / $R^2$}
& \multicolumn{2}{c}{$\rho$ / $R^2$}
& \multicolumn{2}{c}{$\rho$ / $R^2$}
& \multicolumn{2}{c}{$\rho$ / $R^2$}
& \multicolumn{2}{c}{$\rho$ / $R^2$} \\
\midrule

Raw Text &
& 0.504 & -0.571
& 0.570 & -1.277
& 0.528 & -0.729
& 0.462 & -1.107
& 0.493 & -1.879 \\
\midrule

Adjust-Bias & 10-shot
& 0.504 & -0.385
& 0.570 & -0.765
& 0.528 & -0.517
& 0.462 & -0.425
& 0.493 & -1.727 \\

Few-shot & 10-shot
& 0.319 & -1.850
& 0.197 & -1.576
& 0.372 & -0.547
& 0.241 & -0.537
& 0.407 & -0.185 \\

LoRA & 10-shot
& 0.476 & -1.867
& 0.581 & -1.383
& 0.567 & -1.206
& 0.503 & -1.464
& 0.542 & -1.231 \\

Linear-Hidden & 10-shot
& 0.396 & 0.069
& 0.401 & 0.071
& 0.402 & 0.085
& 0.402 & 0.067
& 0.389 & 0.044 \\

Linear-Hidden (GIAA) & 10-shot
& 0.446 & -0.059
& 0.447 & -0.112
& 0.456 & -0.066
& 0.462 & -0.092
& 0.444 & -0.086 \\

Linear-Hidden (Reduce) & 10-shot
& 0.454 & -0.049
& 0.466 & -0.271
& 0.416 & -0.257
& 0.460 & -0.127
& 0.476 & -0.067 \\

\midrule

Adjust-Bias & 100-shot
& 0.504 & -0.310
& 0.570 & -0.672
& 0.528 & -0.441
& 0.462 & -0.321
& 0.493 & -1.562 \\

Few-shot & 100-shot
& {--} & {--}
& {--} & {--}
& {--} & {--}
& 0.254 & -0.533
& {--} & {--} \\

LoRA & 100-shot
& 0.487 & -1.970
& 0.578 & -1.751
& 0.568 & -0.978
& 0.489 & -0.893
& 0.524 & -0.525 \\

Linear-Hidden & 100-shot
& \bfseries 0.604 & 0.363
& \bfseries 0.611 & 0.362
& 0.591 & \bfseries 0.341
& 0.591 & 0.346
& 0.594 & 0.329 \\

Linear-Hidden (GIAA) & 100-shot
& 0.596 & 0.041
& 0.603 & 0.057
& \bfseries 0.596 & 0.043
& 0.584 & -0.014
& \bfseries 0.594 & 0.036 \\

Linear-Hidden (Reduce) & 100-shot
& 0.585 & \bfseries 0.367
& 0.597 & \bfseries 0.382
& 0.558 & 0.322
& \bfseries 0.592 & \bfseries 0.365
& 0.593 & \bfseries 0.373 \\

\bottomrule
\end{tabular}

\caption{Full PIAA results on PARA.}
\label{tab:para_useravg_metrics}
\end{table*}

%% file: tables/lapis_piaa_table.tex
\begin{table*}[t]
\centering
\small
\setlength{\tabcolsep}{3pt}
\renewcommand{\arraystretch}{1.0}

\begin{tabular}{
@{}l l
*{5}{S[table-format=1.3] @{/} S[table-format=-1.3]}
@{}}
\toprule
Method & Support
& \multicolumn{6}{c}{Qwen3-VL}
& \multicolumn{4}{c}{Gemma 3} \\

&
& \multicolumn{2}{c}{2B}
& \multicolumn{2}{c}{4B}
& \multicolumn{2}{c}{8B}
& \multicolumn{2}{c}{4B}
& \multicolumn{2}{c}{12B} \\

&
& \multicolumn{2}{c}{$\rho$ / $R^2$}
& \multicolumn{2}{c}{$\rho$ / $R^2$}
& \multicolumn{2}{c}{$\rho$ / $R^2$}
& \multicolumn{2}{c}{$\rho$ / $R^2$}
& \multicolumn{2}{c}{$\rho$ / $R^2$} \\
\midrule

Raw Text &
& 0.098 & -0.778
& 0.176 & -0.937
& 0.175 & -0.763
& 0.119 & -1.340
& 0.233 & -1.335 \\

\midrule

Adjust-Bias & 10-shot
& 0.098 & -0.399
& 0.176 & -0.382
& 0.175 & -0.345
& 0.119 & -0.284
& 0.233 & -0.587 \\

Few-shot & 10-shot
& 0.142 & -1.265
& 0.221 & -0.380
& 0.264 & -0.480
& 0.127 & -0.354
& 0.227 & -0.459 \\

LoRA & 10-shot
& -0.011 & -0.533
& 0.188 & -1.290
& 0.137 & -1.465
& 0.174 & -1.315
& 0.249 & -1.169 \\

Linear-Hidden & 10-shot
& 0.392 & 0.003
& 0.390 & 0.002
& 0.402 & 0.013
& 0.407 & 0.042
& 0.409 & 0.018 \\

Linear-Hidden (GIAA) & 10-shot
& 0.336 & -0.240
& 0.338 & -0.237
& 0.348 & -0.221
& 0.337 & -0.227
& 0.336 & -0.237 \\

Linear-Hidden (Reduce) & 10-shot
& 0.312 & -0.334
& 0.264 & -0.562
& 0.277 & -0.427
& 0.296 & -0.332
& 0.255 & -0.418 \\

\midrule

Adjust-Bias & 100-shot
& 0.098 & -0.264
& 0.176 & -0.231
& 0.175 & -0.206
& 0.119 & -0.162
& 0.233 & -0.442 \\

Few-shot & 100-shot
& {--} & {--}
& {--} & {--}
& {--} & {--}
& 0.093 & -0.402
& {--} & {--} \\

LoRA & 100-shot
& 0.026 & -0.701
& 0.153 & -1.580
& 0.164 & -1.386
& 0.116 & -0.936
& 0.201 & -1.022 \\

Linear-Hidden & 100-shot
& \bfseries 0.568 & \bfseries 0.321
& \bfseries 0.568 & \bfseries 0.319
& \bfseries 0.573 & \bfseries 0.313
& \bfseries 0.568 & \bfseries 0.328
& \bfseries 0.571 & \bfseries 0.323 \\

Linear-Hidden (GIAA) & 100-shot
& 0.418 & -0.148
& 0.420 & -0.148
& 0.420 & -0.151
& 0.413 & -0.153
& 0.416 & -0.155 \\

Linear-Hidden (Reduce) & 100-shot
& 0.480 & 0.224
& 0.468 & 0.202
& 0.459 & 0.197
& 0.469 & 0.220
& 0.446 & 0.189 \\

\bottomrule
\end{tabular}

\caption{Full PIAA results on LAPIS.}
\label{tab:lapis_useravg_metrics}
\end{table*}

%% file: tables/piaa-vision-encoder.tex
\begin{table*}[t]
\centering
\small
\setlength{\tabcolsep}{4pt}
\renewcommand{\arraystretch}{1.1}
\begin{tabular}{l l c cc}
\toprule
Method & Model & Repr & PARA & LAPIS \\
\midrule
Linear-Hidden & Qwen3-VL 2B & $V_{23}$ 
& \textbf{0.581} / \textbf{0.363} 
& \textbf{0.595} / \textbf{0.360} \\

Linear-Hidden (Reduce) & Qwen3-VL 2B & $V_{23}$ 
& 0.520 / 0.266 
& 0.509 / 0.259 \\

Linear-Hidden & Gemma 3 4B & $V_{26}$ 
& 0.573 / 0.315 
& 0.537 / 0.290 \\

Linear-Hidden (Reduce) & Gemma 3 4B & $V_{26}$ 
& 0.481 / 0.198 
& 0.429 / 0.175 \\
\bottomrule
\end{tabular}
\caption{PIAA performance using vision encoder representations. Values are reported as $\rho / R^2$.}
\label{tab:piaa_vision_encoder}
\end{table*}

%% file: tables/combined_rho_nan_counts.tex
\begin{table*}[t]
\centering
\small
\setlength{\tabcolsep}{3.5pt} %
\renewcommand{\arraystretch}{1.1}

\begin{tabular}{
@{}l l
*{5}{S[table-format=3.0] @{\,(} S[table-format=2.1] @{\,\%)\;}}
@{}}
\toprule
Method & Support
& \multicolumn{6}{c}{Qwen3-VL}
& \multicolumn{4}{c}{Gemma 3} \\

&
& \multicolumn{2}{c}{2B}
& \multicolumn{2}{c}{4B}
& \multicolumn{2}{c}{8B}
& \multicolumn{2}{c}{4B}
& \multicolumn{2}{c}{12B} \\

\midrule
\multicolumn{12}{l}{\textbf{PARA}} \\
\midrule

Few-shot & 10-shot
& 28 & 14.0
& 153 & 76.5
& 42 & 21.0
& 94 & 47.0
& 2 & 1.0 \\

Linear-Hidden & 10-shot
& 4 & 2.0
& 4 & 2.0
& 4 & 2.0
& 4 & 2.0
& 4 & 2.0 \\

Linear-Hidden (Reduce) & 10-shot
& 4 & 2.0
& 4 & 2.0
& 4 & 2.0
& 4 & 2.0
& 4 & 2.0 \\

\midrule

Few-shot & 100-shot
& {--} & {--}
& {--} & {--}
& {--} & {--}
& 71 & 35.5
& {--} & {--} \\

\midrule\midrule
\multicolumn{12}{l}{\textbf{LAPIS}} \\
\midrule

Few-shot & 10-shot
& 46 & 23.0
& 15 & 7.5
& 8 & 4.0
& 120 & 60.0
& 2 & 1.0 \\

LoRA & 10-shot
& 6 & 3.0
& 0 & 0.0
& 0 & 0.0
& 0 & 0.0
& 0 & 0.0 \\

\midrule

Few-shot & 100-shot
& {--} & {--}
& {--} & {--}
& {--} & {--}
& 54 & 27.0
& {--} & {--} \\

\bottomrule
\end{tabular}

\caption{Number of users for which Spearman correlation is undefined due to constant predictions or labels. Rows with zero NaN users for every model are omitted within each dataset block.}
\label{tab:combined_rho_nan_count}
\end{table*}

%% file: tables/para_rho_ci.tex
\begin{table*}[t]
\centering
\small
\setlength{\tabcolsep}{4pt}
\renewcommand{\arraystretch}{1.05}
\begin{tabular}{@{}l l@{\hspace{0.6em}} c c c c c@{}}
\toprule
Method & Support & \multicolumn{3}{c}{Qwen3-VL} & \multicolumn{2}{c}{Gemma 3} \\
& & 2B & 4B & 8B & 4B & 12B \\
& & $[\rho_{\text{min}},\ \rho_{\text{max}}]$
  & $[\rho_{\text{min}},\ \rho_{\text{max}}]$
  & $[\rho_{\text{min}},\ \rho_{\text{max}}]$
  & $[\rho_{\text{min}},\ \rho_{\text{max}}]$
  & $[\rho_{\text{min}},\ \rho_{\text{max}}]$ \\
\midrule

Raw Text &
& \cicell{0.486}{0.521}
& \cicell{0.549}{0.587}
& \cicell{0.510}{0.546}
& \cicell{0.442}{0.481}
& \cicell{0.472}{0.514} \\

\midrule

Few-shot & 10-shot
& \cicell{0.297}{0.339}
& \cicell{0.154}{0.242}
& \cicell{0.341}{0.404}
& \cicell{0.209}{0.272}
& \cicell{0.384}{0.430} \\

\midrule

Adjust-Bias & 100-shot
& \cicell{0.486}{0.522}
& \cicell{0.550}{0.588}
& \cicell{0.510}{0.546}
& \cicell{0.443}{0.481}
& \cicell{0.472}{0.515} \\

LoRA & 100-shot
& \cicell{0.469}{0.505}
& \cicell{0.558}{0.596}
& \cicell{0.548}{0.586}
& \cicell{0.468}{0.510}
& \cicell{0.504}{0.543} \\

Linear-Hidden & 100-shot
& \cicell{0.583}{0.623}
& \cicell{0.591}{0.630}
& \cicell{0.571}{0.611}
& \cicell{0.570}{0.610}
& \cicell{0.572}{0.613} \\

Linear-Hidden (GIAA) & 100-shot
& \cicell{0.577}{0.616}
& \cicell{0.585}{0.622}
& \cicell{0.577}{0.614}
& \cicell{0.565}{0.603}
& \cicell{0.575}{0.613} \\

Linear-Hidden (Reduce) & 100-shot
& \cicell{0.565}{0.604}
& \cicell{0.579}{0.614}
& \cicell{0.538}{0.576}
& \cicell{0.572}{0.610}
& \cicell{0.574}{0.611} \\

\bottomrule
\end{tabular}
\caption{User-averaged Spearman correlation with 95\% bootstrap confidence intervals on PARA.}
\label{tab:para_rho_ci}
\end{table*}

%% file: tables/lapis_rho_ci.tex
\begin{table*}[t]
\centering
\small
\setlength{\tabcolsep}{4pt}
\renewcommand{\arraystretch}{1.05}
\begin{tabular}{@{}l l@{\hspace{0.6em}} c c c c c@{}}
\toprule
Method & Support & \multicolumn{3}{c}{Qwen3-VL} & \multicolumn{2}{c}{Gemma 3} \\
& & 2B & 4B & 8B & 4B & 12B \\
& & $[\rho_{\text{min}},\ \rho_{\text{max}}]$
  & $[\rho_{\text{min}},\ \rho_{\text{max}}]$
  & $[\rho_{\text{min}},\ \rho_{\text{max}}]$
  & $[\rho_{\text{min}},\ \rho_{\text{max}}]$
  & $[\rho_{\text{min}},\ \rho_{\text{max}}]$ \\
\midrule

Raw Text &
& \cicell{0.077}{0.118}
& \cicell{0.153}{0.199}
& \cicell{0.150}{0.200}
& \cicell{0.098}{0.139}
& \cicell{0.204}{0.263} \\

\midrule

Few-shot & 10-shot
& \cicell{0.113}{0.170}
& \cicell{0.193}{0.251}
& \cicell{0.235}{0.294}
& \cicell{0.093}{0.161}
& \cicell{0.192}{0.259} \\

\midrule

Adjust-Bias & 100-shot
& \cicell{0.076}{0.118}
& \cicell{0.152}{0.200}
& \cicell{0.149}{0.202}
& \cicell{0.098}{0.139}
& \cicell{0.204}{0.263} \\

LoRA & 100-shot
& \cicell{-0.001}{0.050}
& \cicell{0.131}{0.176}
& \cicell{0.135}{0.192}
& \cicell{0.094}{0.138}
& \cicell{0.171}{0.230} \\

Linear-Hidden & 100-shot
& \cicell{0.542}{0.593}
& \cicell{0.544}{0.593}
& \cicell{0.547}{0.597}
& \cicell{0.542}{0.595}
& \cicell{0.546}{0.596} \\

Linear-Hidden (GIAA) & 100-shot
& \cicell{0.379}{0.458}
& \cicell{0.382}{0.460}
& \cicell{0.383}{0.458}
& \cicell{0.374}{0.452}
& \cicell{0.377}{0.454} \\

Linear-Hidden (Reduce) & 100-shot
& \cicell{0.455}{0.506}
& \cicell{0.442}{0.495}
& \cicell{0.430}{0.487}
& \cicell{0.442}{0.496}
& \cicell{0.420}{0.471} \\

\bottomrule
\end{tabular}

\caption{User-averaged Spearman correlation with 95\% bootstrap confidence intervals on LAPIS.}
\label{tab:lapis_rho_ci}
\end{table*}

%% file: figures/sample_aug.tex
\begin{figure}[t]
  \centering
    \includegraphics[width=\linewidth]{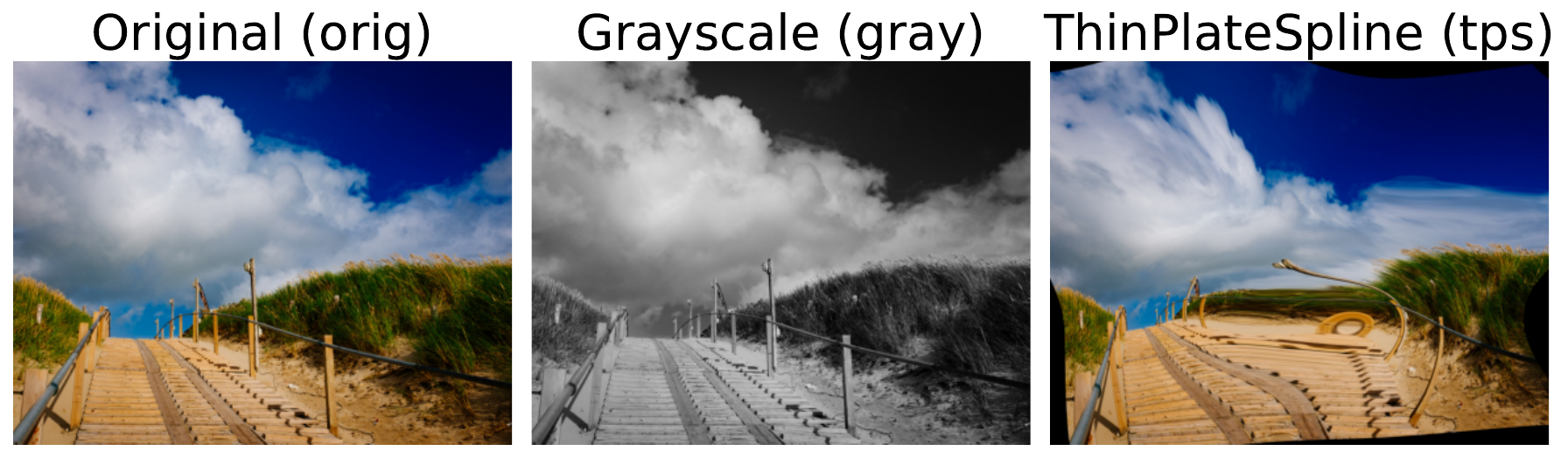}
  \caption{Example images from AADB with applied augmentations.}
  \label{fig:sample-aug}
\end{figure}

%% file: figures/probing_aug_attr.tex
\begin{figure}[t]
  \centering
  \begin{minipage}[t]{0.5\linewidth}
    \centering
    \includegraphics[width=\linewidth]{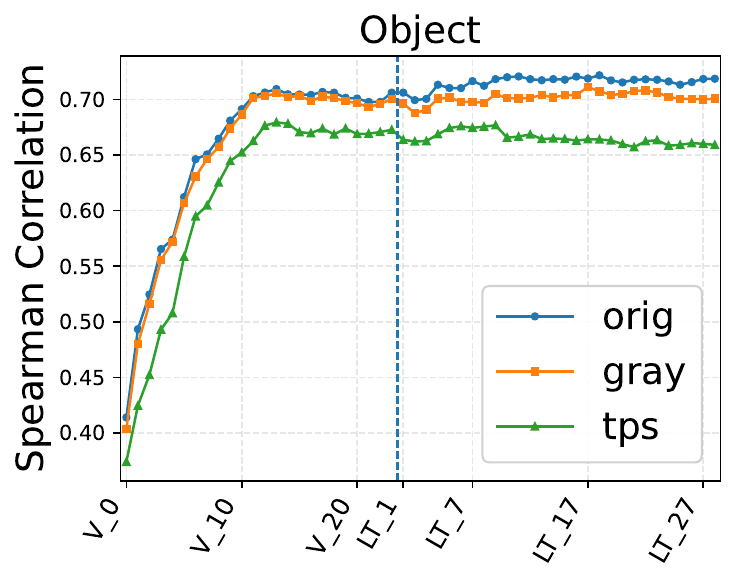}
  \end{minipage}%
  \begin{minipage}[t]{0.5\linewidth}
    \centering
    \includegraphics[width=\linewidth]{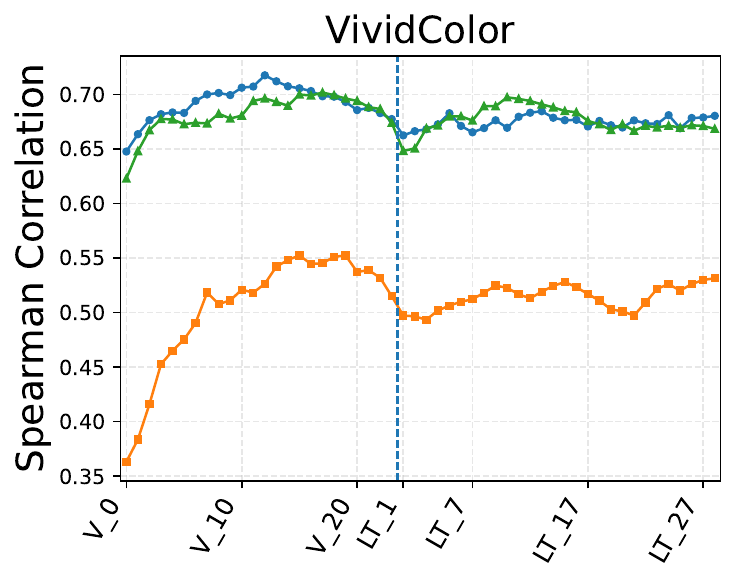}
  \end{minipage}%
  \caption{Probing performance under different image augmentations for Qwen3-VL 2B.}
  \label{fig:probing-aug-attr}
\end{figure}

%% file: tables/probing_table_para.tex
\begin{table*}[t]
\centering
\small
\setlength{\tabcolsep}{3pt}
\renewcommand{\arraystretch}{1.0}

\begin{tabular}{
l
S @{\,(} c @{)\hspace{1em}}
S @{\,(} c @{)\hspace{1em}}
S @{\,(} c @{)\hspace{1em}}
S @{\,(} c @{)\hspace{1em}}
S @{\,(} c @{)\hspace{1em}}
S @{\,(} c @{)\hspace{1em}}
S @{\,(} c @{)}
}
\toprule

Attribute
& \multicolumn{6}{c}{Qwen3-VL}
& \multicolumn{4}{c}{Gemma 3}
& \multicolumn{4}{c}{DINOv3} \\

& \multicolumn{2}{c}{2B}
& \multicolumn{2}{c}{4B}
& \multicolumn{2}{c}{8B}
& \multicolumn{2}{c}{4B}
& \multicolumn{2}{c}{12B}
& \multicolumn{2}{c}{ViT-B/16}
& \multicolumn{2}{c}{ViT-L/16} \\

\midrule

color & 0.884 & 13 & \bfseries 0.886 & 18 & 0.886 & 17 & 0.874 & 8  & 0.880 & 13 & 0.845 & 8  & 0.855 & 16 \\
composition & 0.887 & 12 & \bfseries 0.890 & 19 & 0.889 & 13 & 0.880 & 10 & 0.887 & 16 & 0.845 & 8  & 0.855 & 17 \\
content & 0.871 & 13 & \bfseries 0.874 & 17 & 0.872 & 17 & 0.863 & 9  & 0.867 & 13 & 0.818 & 8  & 0.832 & 17 \\
dof & 0.883 & 13 & \bfseries 0.887 & 19 & 0.886 & 19 & 0.874 & 9  & 0.879 & 15 & 0.838 & 8  & 0.847 & 17 \\
light & 0.869 & 13 & \bfseries 0.876 & 18 & 0.875 & 18 & 0.861 & 10 & 0.866 & 14 & 0.828 & 8  & 0.839 & 17 \\
quality & 0.905 & 15 & \bfseries 0.910 & 17 & 0.909 & 19 & 0.893 & 7  & 0.899 & 15 & 0.853 & 8  & 0.863 & 17 \\
Overall Score & 0.910 & 15 & \bfseries 0.913 & 18 & 0.913 & 18 & 0.900 & 9  & 0.906 & 15 & 0.861 & 8  & 0.872 & 17 \\

\bottomrule
\end{tabular}

\caption{Highest Spearman correlation achieved by probing on $\mathbf{LT}$ layers for PARA.
Layer indices are shown in parentheses.}
\label{tab:probing_best_layers_para}
\end{table*}

%% file: figures/probing-para.tex
\begin{figure}[t]
  \centering
    \includegraphics[width=\columnwidth]{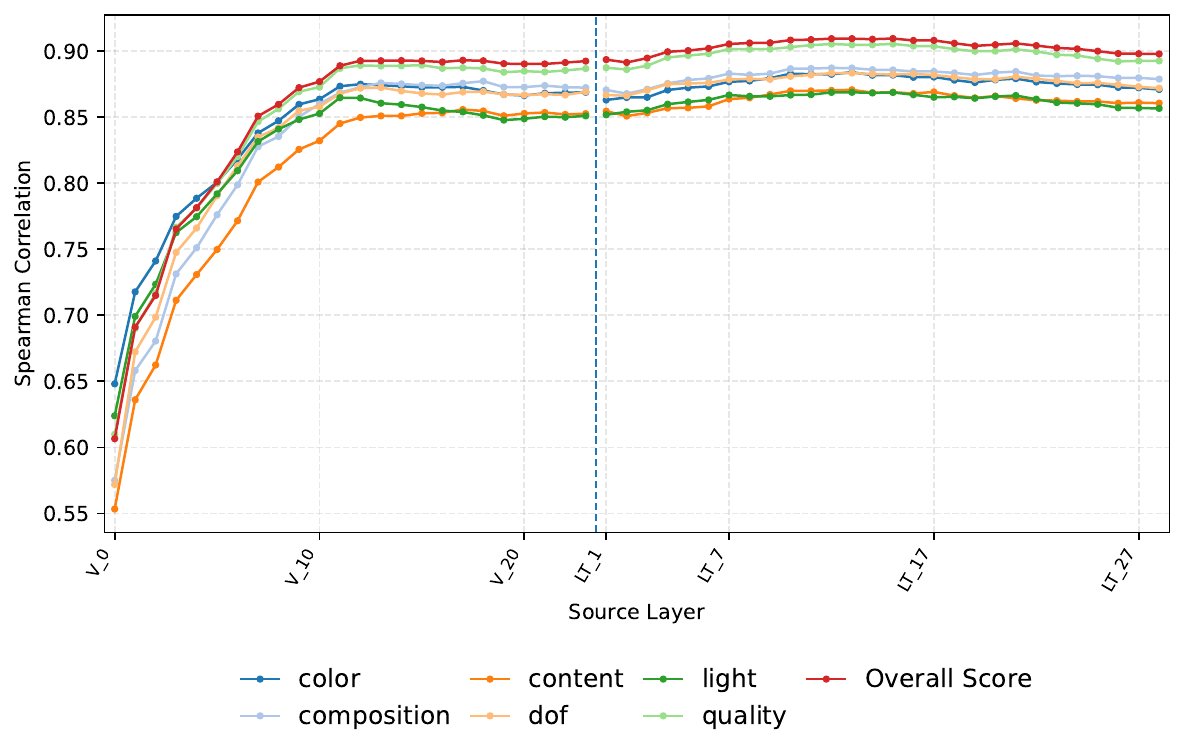}
  \caption{Layer-wise probing performance across $\mathbf{V}$ and $\mathbf{LT}$ layers for Qwen3-VL 2B on PARA.}
  \label{fig:probing-para}
\end{figure}

%% file: tables/probing_prompts_table.tex
\begin{table*}[t]
\centering
\small
\setlength{\tabcolsep}{3pt}
\renewcommand{\arraystretch}{1.0}

\begin{tabular}{
l
S @{\,(} c @{)\hspace{1em}}
S @{\,(} c @{)\hspace{1em}}
S @{\,(} c @{)\hspace{1em}}
S @{\,(} c @{)}
}
\toprule

Attribute
& \multicolumn{2}{c}{Base}
& \multicolumn{2}{c}{Numeric}
& \multicolumn{2}{c}{Attribute}
& \multicolumn{2}{c}{Unrelated} \\

\midrule

BalancingElements & \bfseries 0.325 & 13 & 0.323 & 10 & 0.323 & 11 & 0.319 & 0 \\
ColorHarmony      & 0.516 & 9  & 0.510 & 12 & 0.518 & 16 & \bfseries 0.531 & 11 \\
Content           & 0.633 & 10 & \bfseries 0.634 & 12 & 0.629 & 10 & 0.626 & 9 \\
DoF               & 0.535 & 10 & \bfseries 0.541 & 11 & 0.531 & 21 & 0.535 & 9 \\
Light             & \bfseries 0.509 & 14 & 0.505 & 19 & 0.508 & 17 & 0.491 & 13 \\
MotionBlur        & 0.165 & 12 & 0.142 & 11 & \bfseries 0.176 & 12 & 0.167 & 10 \\
Object            & 0.722 & 18 & 0.723 & 11 & \bfseries 0.727 & 11 & 0.726 & 24 \\
Repetition        & 0.461 & 3  & 0.463 & 3  & 0.456 & 4  & \bfseries 0.463 & 4 \\
RuleOfThirds      & 0.288 & 11 & 0.286 & 11 & \bfseries 0.295 & 13 & 0.288 & 10 \\
Symmetry          & 0.315 & 10 & 0.312 & 0  & 0.314 & 6  & \bfseries 0.331 & 12 \\
VividColor        & 0.686 & 0  & 0.688 & 5  & \bfseries 0.696 & 23 & 0.688 & 28 \\
score             & \bfseries 0.725 & 5  & 0.721 & 12 & 0.724 & 25 & 0.718 & 7 \\

\bottomrule
\end{tabular}

\caption{Probing performance on Qwen3-VL 2B under different prompt formulations.}
\label{tab:probing_best_by_prompt}
\end{table*}

%% file: tables/para_resize_table.tex
\begin{table}[t]
\centering
\small
\setlength{\tabcolsep}{3pt}
\renewcommand{\arraystretch}{1.0}
\begin{tabular}{l l c}
\toprule
Model & Method & $\rho$ / $R^2$ \\
\midrule
Qwen3-VL 2B & Raw Text (Original)      & 0.504 / -0.571 \\
           & Raw Text (Resized)       & 0.505 / -0.511 \\
           & LoRA (Resized, 100-shot) & 0.487 / -1.970 \\
\midrule
Qwen3-VL 4B & Raw Text (Original)      & 0.570 / -1.277 \\
           & Raw Text (Resized)       & 0.568 / -1.117 \\
           & LoRA (Resized, 100-shot) & 0.578 / -1.751 \\
\midrule
Qwen3-VL 8B & Raw Text (Original)      & 0.528 / -0.729 \\
           & Raw Text (Resized)       & 0.532 / -0.724 \\
           & LoRA (Resized, 100-shot) & 0.568 / -0.978 \\
\bottomrule
\end{tabular}
\caption{Effect of image resizing on PIAA performance for Qwen3-VL models on PARA.}
\label{tab:para_resize_table}
\end{table}

%% file: figures/combined_heatmaps.tex
\begin{figure}[t]
  \centering
    \includegraphics[width=\linewidth]{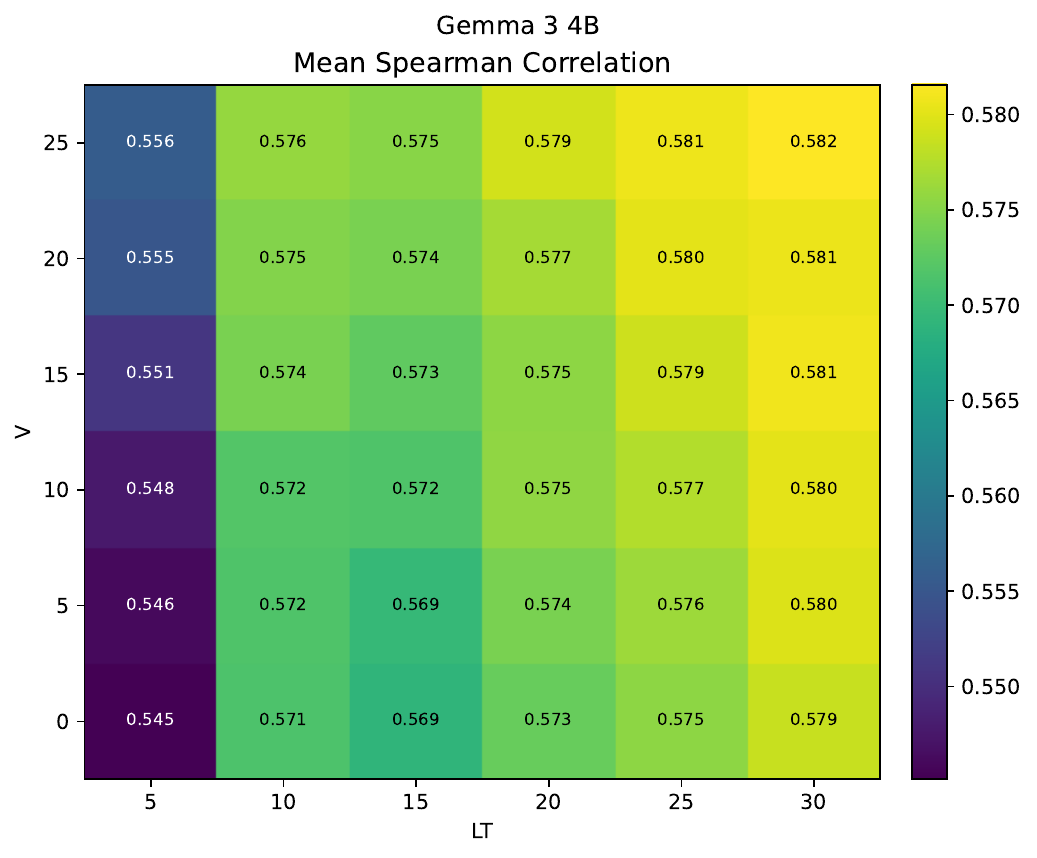}
    \includegraphics[width=\linewidth]{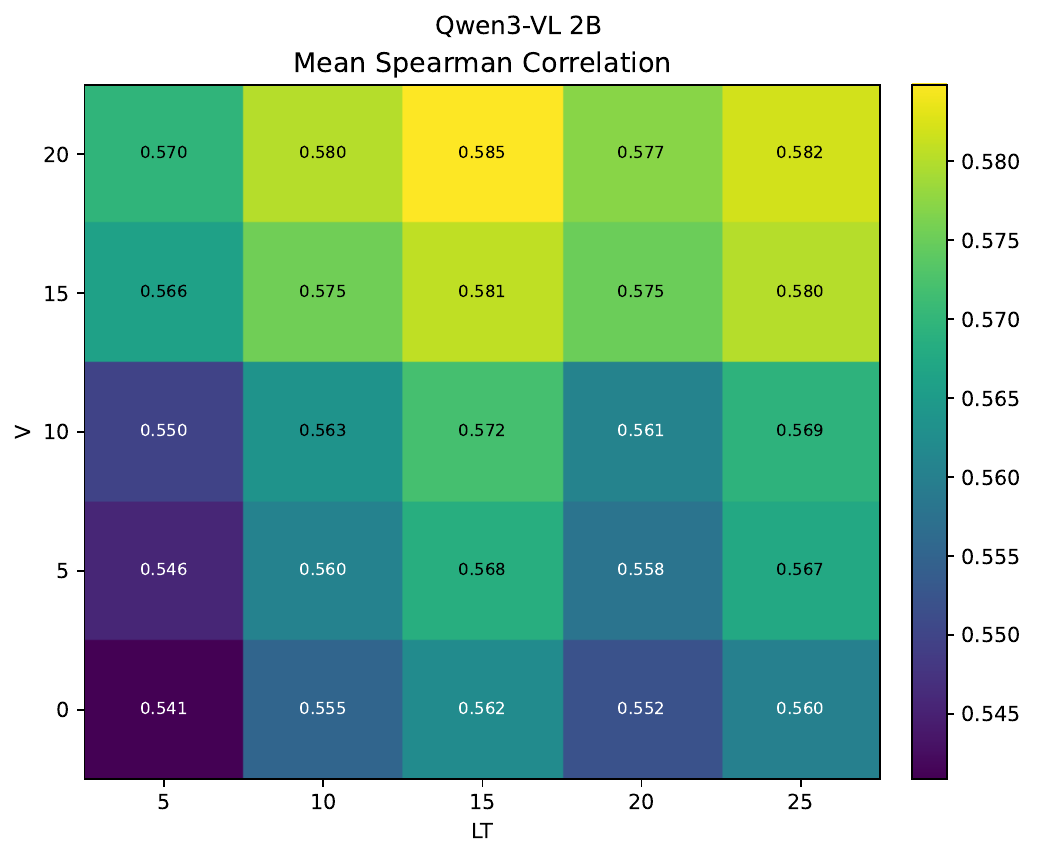}

  \caption{User-averaged Spearman correlation obtained by Linear-Hidden PIAA using combined representations from the vision encoder ($\mathbf{V}_i$) and language decoder ($\mathbf{LT}_j$).}
  \label{fig:combined_heatmaps}
\end{figure}

%% file: tables/logo-holdout-summary.tex
\begin{table*}[t]
\centering
\small
\setlength{\tabcolsep}{4pt}
\renewcommand{\arraystretch}{1.1}
\begin{tabular}{lcc}
\toprule
Train / Test Split & Qwen3-VL 2B & Gemma 3 4B \\
\midrule
Train=FIGURATIVE, Test=ABSTRACT & 0.183 / -0.596 & 0.167 / -0.697 \\
Train=ABSTRACT, Test=FIGURATIVE & 0.145 / -1.072 & 0.134 / -1.062 \\
\bottomrule
\end{tabular}
\caption{User-averaged Spearman correlation and $R^2$ of Linear-Hidden on LAPIS under cross-style evaluation.}
\label{tab:logo-holdout-summary}
\end{table*}